\documentclass{article}

\PassOptionsToPackage{numbers,compress}{natbib}
\usepackage[preprint]{neurips_2026}

\usepackage[utf8]{inputenc} % allow utf-8 input
\usepackage[T1]{fontenc}    % use 8-bit T1 fonts
\usepackage{url}            % simple URL typesetting
\usepackage{amsfonts}       % blackboard math symbols
\usepackage{nicefrac}       % compact symbols for 1/2, etc.
\usepackage{microtype}      % microtypography
\usepackage{xcolor}         % colors
\usepackage{amsmath}
\usepackage{graphicx}
\usepackage{tabularx}
\usepackage{multirow,booktabs}
\usepackage{xcolor}    % 颜色支持
\usepackage{colortbl}  % 表格上色（必须）
\usepackage{arydshln}
\usepackage{bbding}
\usepackage{caption}
\usepackage{ulem}
\usepackage{wrapfig}
\usepackage{pifont}
\usepackage{titletoc}
\usepackage[most]{tcolorbox}
\usepackage{fontawesome5}

\definecolor{mycitecolor}{RGB}{47, 129, 240}
\definecolor{myurlcolor}{RGB}{47, 129, 240}
\definecolor{mylinkcolor}{RGB}{255, 92, 59}

\usepackage[
    colorlinks=true,
    citecolor=mycitecolor,  % 参考文献引用颜色
    linkcolor=mylinkcolor, % 内部链接颜色
    urlcolor=myurlcolor,  % 网址颜色
]{hyperref}

\newcolumntype{Y}{>{\centering\arraybackslash}X}

\newcommand{\Figure}{\textcolor{mylinkcolor}{Figure}}
\newcommand{\Table}{\textcolor{mylinkcolor}{Table}}
\newcommand{\Eq}{\textcolor{mylinkcolor}{Eq.}}
\newcommand{\Sec}{\textcolor{mylinkcolor}{\S}}
\newcommand{\Appendix}{\textcolor{mylinkcolor}{Appendix}}
\newcommand{\eg}{\textit{e.g.}}
\newcommand{\ie}{\textit{i.e.}}
\newcommand{\cf}{\textit{cf.}}
\newcommand{\prompt}[1]{%
\begin{tcolorbox}[
    enhanced,
    colback=mycitecolor!8!white,
    colframe=mycitecolor,
    leftrule=2mm,
    rightrule=0mm,
    toprule=0mm,
    bottomrule=0mm,
    arc=0mm,
    left=5pt,
    right=5pt,
    top=5pt,
    bottom=5pt,
    breakable,
    leftlower=2mm,
    leftupper=2mm
]
\normalsize 
#1
\end{tcolorbox}
}

\title{Efficient SAM3 Adaptation for Multi-Class Semantic Segmentation via Dense Competitive Representations}

\author{%
    Wenbin Liao\textsuperscript{1,2}\hspace{1em}
    Hao Zhu\textsuperscript{1,2}\hspace{1em}
    Yike Ma\textsuperscript{1}\hspace{1em}
    Hao Jiang\textsuperscript{1}\hspace{1em}
    Feng Dai\textsuperscript{1,}\thanks{Corresponding Author} \\
    \textsuperscript{1}Institute of Computing Technology, Chinese Academy of Science \\
    \textsuperscript{2}School of Computer Science and Technology, University of Chinese Academy of Sciences \\
    \texttt{\{liaowenbin23z, zhuhao22z, ykma, jianghao, fdai\}@ict.ac.cn}
}

\begin{document}

\maketitle

\begin{abstract}
The Segment Anything Model 3 (SAM3) has achieved significant progress in \textit{Promptable Concept Segmentation (PCS)} by processing short noun phrases to generate segmentation masks with unique instance identifiers. 
% While SAM3 exhibits strong performance in single-concept segmentation, complex real-world scenarios typically require the simultaneous segmentation of multiple concepts. 
% Currently, SAM3 performs independent inference for each concept and subsequently aggregates the predicted masks. 
While SAM3 excels at single-concept segmentation, it falls short in complex, real-world environments where multiple concepts must be segmented simultaneously.
% In particular, SAM3 struggles to differentiate conceptually similar categories, frequently producing ``double-high'' responses for identical pixel regions. 
In particular, SAM3 typically fails to simultaneously process concepts with semantic overlap, leading to semantic misclassifications and redundant mask predictions.
% This ambiguity inevitably results in semantic misclassifications and overlapping masks. 
Moreover, as the number of concepts increases, the independent inference paradigm introduces prohibitive inference latency. 
To address these limitations, we propose the \textbf{D}ense \textbf{C}ompetitive \textbf{R}epresentations \textbf{SAM} (\textbf{DCR-SAM}) framework. By decoupling cross-modal interactions, the proposed method alleviates the inference latency. To resolve semantic misclassifications, DCR-SAM dynamically aligns textual concepts with visual representations and incorporates a dense competition head to enforce explicit inter-class competition. Furthermore, a learnable background token is applied to absorb non-target objects dynamically. These mechanisms effectively suppress overlapping predictions and generate high-fidelity masks. Extensive experimental evaluations across multiple benchmarks demonstrate that the proposed framework: \ding{182} achieves a 2.3\% $\sim$ 10.7\% performance improvement over state-of-the-art methods, \ding{183} delivers a 33$\times$ inference speedup compared to the native SAM3, and \ding{184} exhibits strong robustness across diverse scenarios. \href{https://github.com/set-path/DCR-SAM}{\textcolor[RGB]{153,0,0}{Our code is publicly available at GitHub \faGithub}}.
\end{abstract}

\section{Introduction}

The evolution of Visual Foundation Models (VFMs)~\citep{dino,dinov2,dinov3,sam,sam2,sam3} achieved a breakthrough with the Segment Anything Model 3~\citep{sam3} (SAM3). As a pioneering architecture in \textit{Promptable Concept Segmentation (PCS)}, SAM3~\citep{sam3} demonstrates exceptional capability in single-concept segmentation guided by short noun phrases (\eg, a red billboard). By incorporating deep cross-modal attention mechanisms, the model leverages textual prompts to dynamically condition visual representations, enabling precise mask generation.

While SAM3~\citep{sam3} excels at single-concept segmentation, complex real-world scenarios~\citep{segearthov3,lin2026progressive,bundele2025memory} (\eg, autonomous driving and remote sensing) require simultaneous identification and segmentation of multiple concepts within a single scene. Conventional semantic segmentation methods~\citep{ocrnet,deeplabv3,segman,EoMT} achieve multi-class segmentation by assigning individual pixels to a set of predefined classes. However, these methods face limitations in complex environments. They typically rely on narrow, task-specific feature representations and lack broad semantic priors of large-scale foundation models. 
% 这一段需要改一下，可以直接就写真实世界场景通常需要同时分割多个概念，因此现有方法如何做，然后存在哪些问题（没有大规模预训练的prior）。On the other hand, SAM3可以在单一概念下做的很好，但是在这种多类场景下不好，接着直接引出来两个问题。
% This deficiency restricts their discriminative capacity and feature robustness in challenging scenes. Given the robust visual-linguistic representations acquired during SAM3~\citep{sam3} pre-training, adapting this foundation model for multi-class segmentation presents a promising approach.
On the other hand, SAM3~\citep{sam3} leverage robust visual-linguistic representations to achieve remarkable performance in single-concept segmentation. However, it faces two critical challenges when extended to multi-class scenarios: \ding{182} SAM3 typically fails to simultaneously process similar concepts, leading to semantic misclassifications and redundant mask predictions. \ding{183} As the number of target concepts increases, the original SAM3 inference paradigm incurs exceptionally high inference latency.

\begin{figure}[t]
    \centering
    
    % --- 第一行：放置图片，使用 [c] 参数使其垂直居中对齐 ---
    \begin{minipage}[c]{0.5\textwidth}
        \centering
        % 替换为你的第一张图片文件名，可按需调整 width
        \includegraphics[width=\linewidth]{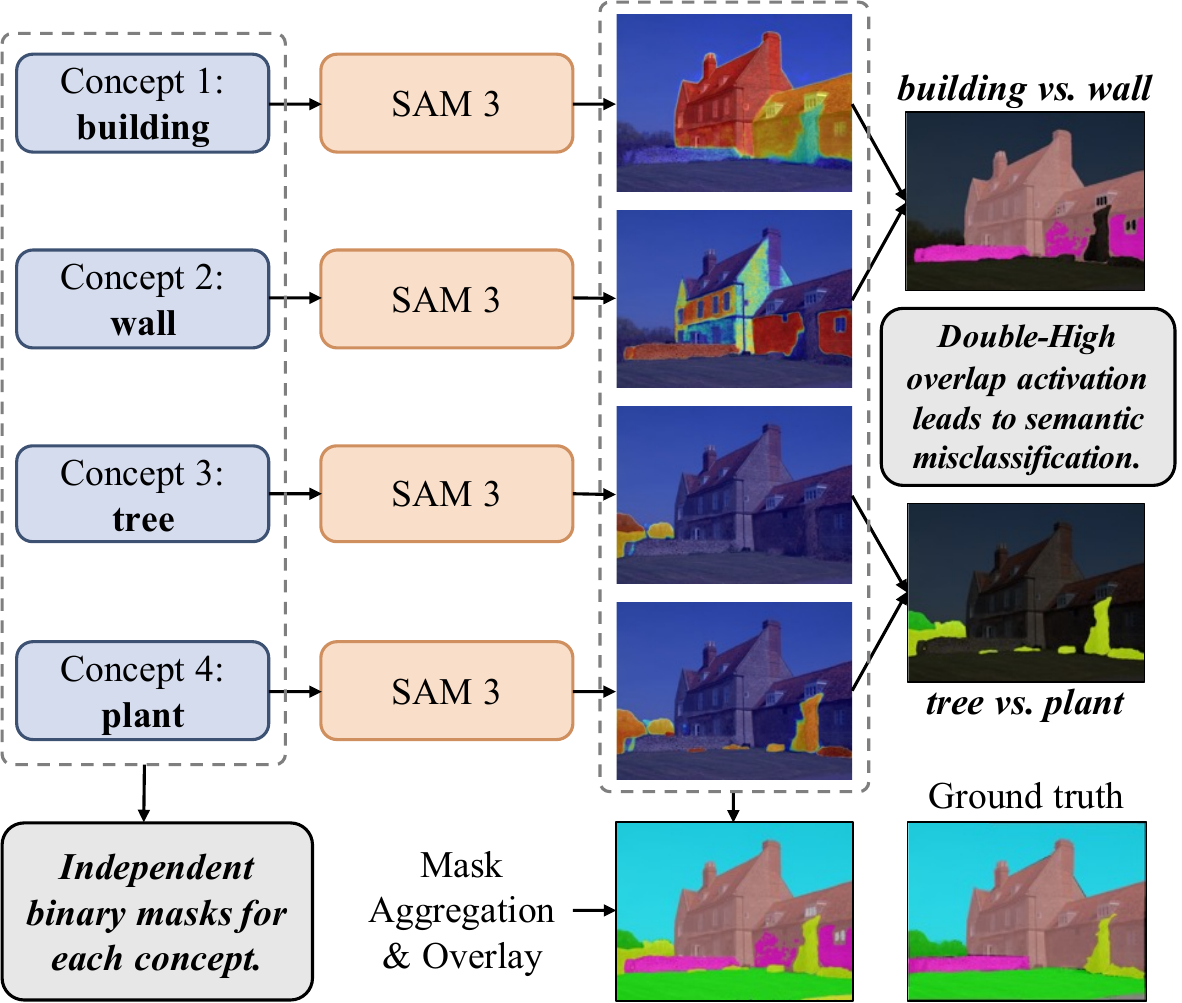}
    \end{minipage}\hfill
    \begin{minipage}[c]{0.46\textwidth}
        \centering
        % 替换为你的第二张图片文件名
        \includegraphics[width=\linewidth]{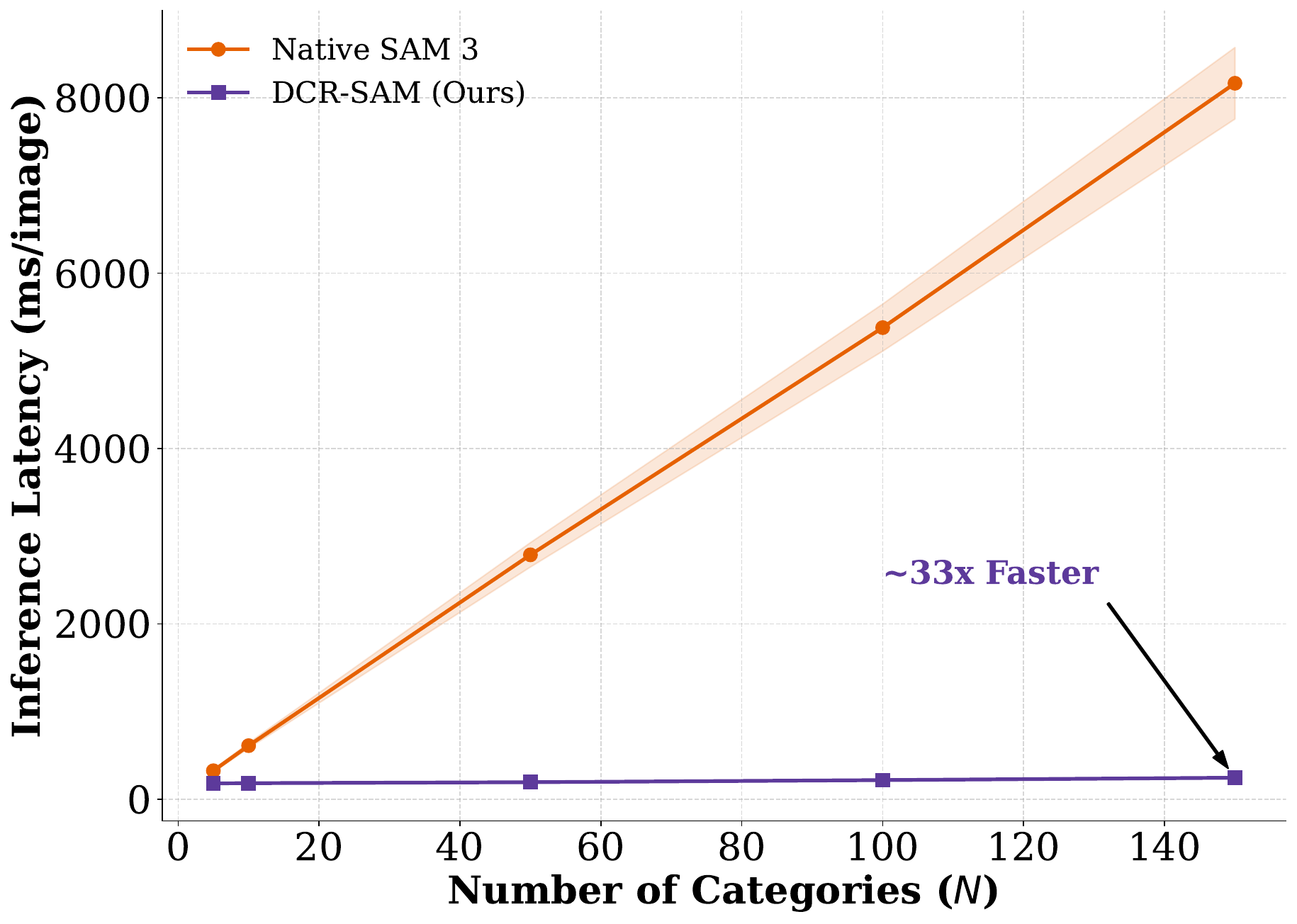}
    \end{minipage}
    
    \vspace{0.8em} % 调整图片与图注之间的垂直间距
    
    % --- 第二行：放置图注，使用 [t] 参数使其顶部（首行）对齐 ---
    \begin{minipage}[t]{0.5\textwidth}
        \caption{\textbf{Limitations of direct mask aggregation in SAM3.} Treating multi-class segmentation as isolated binary segmentation tasks lacks inter-class competition, causing ``double-high'' activation for similar concepts and leading to severe semantic misclassifications in the final mask.}
        \label{fig:limitation}
    \end{minipage}\hfill
    \begin{minipage}[t]{0.46\textwidth}
        \caption{\textbf{Comparison of inference latency.} The inference latency of the original SAM3 increases linearly $\mathcal{O}(n)$ with the number of categories, whereas our method achieves a 33x speedup.}
        \label{fig:efficiency_comparison}
    \end{minipage}
    
\end{figure}

Recent works~\citep{segearthov3,coco-sam3} achieve multi-class segmentation by aggregating the individual masks independently predicted by SAM3 for multiple categories. However, this paradigm inherently treats the multi-class segmentation as a series of independent binary segmentation tasks. Consequently, there is a complete absence of competition among categories, which leaves the model without explicit constraints to differentiate semantically similar concepts. As shown in \Figure~\ref{fig:limitation}, the model frequently produces ``double-high'' activations for identical pixel regions when processing similar concepts (\eg, tree \textit{vs.} plant). This unconstrained behavior inevitably leads to semantic misclassifications and overlapping masks, severely degrading the reliability of the dense prediction results.

On the other hand, the standard architecture of SAM3 requires executing independent and computationally expensive cross-attention operations for each concept to acquire concept-specific dense visual features, followed by an independent decoding process. As shown in \Figure~\ref{fig:efficiency_comparison}, our inference latency scaling analysis reveals that on the ADE20K~\citep{ade20k} dataset with 150 categories, the inference time scales linearly ($\mathcal{O}(n)$) as the number of categories increases, ultimately resulting in severe inference latency.

To address these limitations, we propose the \textbf{D}ense \textbf{C}ompetitive \textbf{R}epresentations \textbf{SAM} (\textbf{DCR-SAM}) framework, which adapts SAM3 into an efficient architecture for multi-class segmentation. Specifically, the proposed framework resolves the aforementioned challenges through three key designs: \ding{182} To overcome the $\mathcal{O}(n)$ inference latency, we decouple the deep cross-modal interactions by replacing class-specific text queries with a unified learnable prompt sequence. This approach enables the extraction of a globally contextualized visual representation shared across all concepts through a single forward pass, achieving a nearly 33$\times$ inference speedup compared to the native SAM3, as shown in \Figure~\ref{fig:efficiency_comparison}. \ding{183} To resolve semantic misclassifications among similar concepts, we transition from isolated decoding to a dense competitive paradigm through the introduction of the \textit{Vision-Conditioned Text Refinement (VCTR)} and \textit{Vision-Guided Text Pooling (VGTP)} modules, coupled with a dense competition head. \textit{VCTR} explicitly aligns textual sequence with the modified visual representation by injecting image-specific visual cues, whereas \textit{VGTP} dynamically compresses this aligned sequence into highly discriminative global class anchors. By enforcing explicit inter-class competition, the proposed method effectively alleviates ``double-high'' activations and generates high-fidelity masks. \ding{184} Furthermore, to enhance adaptability across diverse unconstrained scenarios, we incorporate a dynamically learnable background token to absorb non-target objects adaptively.

The main contributions of our work are summarized as follows:
\begin{itemize}
    \item We reveal how the independent inference paradigm of SAM3 induces severe semantic misclassification among conceptually similar categories and results in an $\mathcal{O}(n)$ inference latency.
    \item We propose \textbf{DCR-SAM}, by decoupling cross-modal interactions to alleviate inference latency and dynamically aligning textual concepts with visual representation, the proposed method effectively suppresses overlapping predictions to generate high-fidelity masks.
    \item Extensive experimental evaluations demonstrate that our method achieves state-of-the-art (SOTA) performance across multiple domain benchmarks while exhibiting excellent inference efficiency.
\end{itemize}

\section{Related Work}

\paragraph{Promptable Segmentation Models}
The Segment Anything Model (SAM)~\citep{sam,sam2,sam3} represents a significant milestone for promptable segmentation models, demonstrating remarkable cross-domain generalization across diverse downstream applications~\citep{segearthov3,blushtein2025effectiveness,tang2026cad,niehorster2026eye,bundele2025memory,lin2026progressive}. Central to this success is the \textit{Promptable Visual Segmentation (PVS)} paradigm introduced by SAM~\citep{sam} and SAM 2~\citep{sam2}, where visual prompts (\eg, points, boxes, or masks) guide the segmentation of specific targets. Driven by large-scale, high-quality datasets, this paradigm establishes a robust baseline for generalized segmentation. Building upon this foundation, SAM3~\cite{sam3} extends the architecture from \textit{PVS} to \textit{Promptable Concept Segmentation (PCS)}, enabling the model to segment all instances of a given concept using short noun phrases (\eg, a red billboard). Recent studies~\citep{sam3-i,concept-bank,efficientsam3,sam3-litetext,sam3-adapter} explore improvements to this architecture. For example, EfficientSAM3~\citep{efficientsam3,sam3-litetext} enhances the overall inference efficiency of SAM3 through distillation, SAM3-Adapter~\citep{sam3-adapter} injects prior heuristic information for downstream tasks via a lightweight staged adapter. However, these works fundamentally retain the inherent serial inference mechanism of the native architecture. Consequently, they incur non-negligible inference latency and semantic misclassification when applied to real-world multi-class segmentation tasks. Meanwhile CoCo-SAM3~\citep{coco-sam3} proposes a training-free framework that separates the inference process into intra-class enhancement and inter-class competition modules to alleviate semantic misclassification. Nevertheless, this method relies heavily on complex score calibration and multi-prompt encoding. In this work, we alleviate these issues at the architectural level by decoupling cross-modal interactions. Furthermore, by dynamically aligning textual anchors with visual representations and introducing explicit inter-class competition, the proposed framework effectively mitigates semantic misclassification and generates high-fidelity masks.

\paragraph{Multi-Class Semantic Segmentation and Foundation Models}
Multi-class semantic segmentation is a fundamental task in computer vision, with the objective of assigning a specific semantic label to every pixel within an image. The field has witnessed remarkable progress, evolving from early CNN-based models, such as FCN~\citep{fcn} and U-Net~\citep{unet}, to modern Transformer-based architectures, such as SegFormer~\cite{segformer}, Mask2Former~\citep{mask2former} and EoMT~\citep{EoMT}. While these methods excel at efficient dense prediction, they typically rely on narrow, task-specific feature representations, thereby lacking the broad semantic priors inherent in large-scale foundation models. Several vision-language models, such as CLIP~\citep{clip}, SigLIP~\cite{siglip}, and ALIGN~\citep{align}, emphasize global image-text alignment and inherently lack pixel-level granularity, rendering them unsuitable for direct application to downstream dense prediction tasks. Conversely, self-supervised vision foundation models, such as DINO~\citep{dino,dinov2,dinov3}, capture excellent fine-grained pixel-level representations on massive image corpora, yet they lack explicit cross-modal semantic guidance and require the training of task-specific decoder heads for multi-class segmentation. In this work, we fully leverage the powerful cross-modal priors of SAM3~\citep{sam3} to enhance the potential of the model in dense prediction tasks.

\section{Methodology}

\begin{figure}
  \centering
  \includegraphics[width=\textwidth]{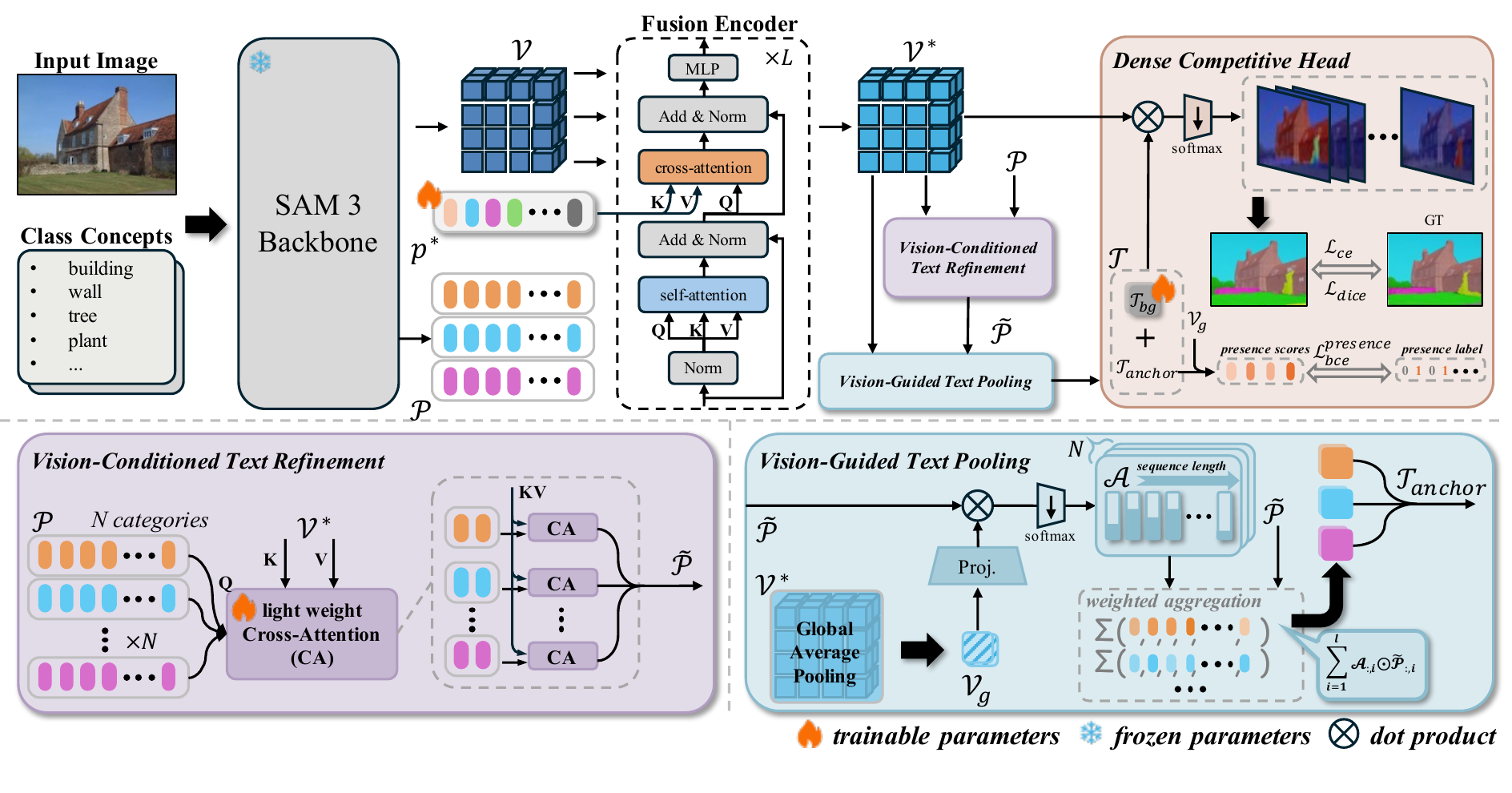}
  \caption{\textbf{Overview of our proposed DCR-SAM framework.} It transforms SAM3 from an independent decoding paradigm to a dense competitive paradigm. First, we alleviate the inference latency via Decoupled Visual Conditioning to obtain a globally contextualized visual representation. Then, we align text and visual features using the \textit{VCTR} and \textit{VGTP} modules. Finally, we design a competition head to enforce explicit inter-class competition and generate high-fidelity masks.}
  \label{fig:architecture}
\end{figure}

In this section, we first briefly review the \textit{Promptable Concept Segmentation (PCS)} pipeline of SAM3 (\Sec\ref{method:review_sam3}). Subsequently, we elaborate on the overall architecture of our proposed DCR-SAM framework (\Sec\ref{method:dcr_sam}). The overview of our method is illustrated in \Figure~\ref{fig:architecture}. Finally, we introduce the joint optimization and the corresponding objective functions (\Sec\ref{method:optimization}).

\subsection{Preliminary}
\label{method:review_sam3}

Unlike dual-encoder architectures such as CLIP~\citep{clip}, SAM3~\citep{sam3} employs an encoder-decoder architecture where the core modules comprise the vision encoder, the text encoder, the fusion encoder, and the semantic decoding head (for simplicity, we exclusively discuss the modules relevant to this work). Given an input image $I \in \mathbb{R}^{H \times W \times 3}$ and a target concept set comprising $N$ categories $C = \{c_1, c_2, \dots, c_N\}$, the network initially processes the image through a vision encoder to extract the unconditioned visual embeddings $\mathcal{V} \in \mathbb{R}^{hw \times d}$, where $d$ denotes the embedding dimension, $h$ and $w$ depend on the patch resolution of the vision encoder. Simultaneously, each specific concept $c_i$ is encoded into prompt tokens $p_{i} \in \mathbb{R}^{l \times d}$ via a text encoder, where $l$ denotes the sequence length. The complete set of prompt tokens is denoted as $\mathcal{P}=\{p_1,p_2,\dots,p_N\}$. Subsequently, to achieve concept-specific localization, the unconditioned visual embeddings must be dynamically conditioned on the prompt tokens. Following the native architecture of SAM3~\citep{sam3}, a fusion encoder $E_{fusion}$ accepts the unconditioned visual embeddings $\mathcal{V}$ and conditions these embeddings on the prompt tokens $p_i$ using a stack of Transformer~\citep{attention} blocks equipped with self-attention and cross-attention layers. The output of fusion encoder constitutes the conditioned visual embeddings:
\begin{equation}
    \mathcal{V}_{cond}^{i} = E_{fusion}(\mathcal{V}, p_{i}),\quad \mathcal{V}_{cond}^{i} \in \mathbb{R}^{hw \times d}.
    \label{eq:fusion}
\end{equation}
Finally, these conditioned embeddings are fed into a semantic decoding head, which is adapted from the MaskFormer~\citep{maskformer} architecture, to decode the binary semantic logits map $\mathcal{M}_{i}$ for the specific concept $c_i$:
\begin{equation}
    \mathcal{M}_{i} = \mathrm{SegHead}(V_{cond}^{i}),\quad \mathcal{M}_{i} \in \mathbb{R}^{h \times w}.
    \label{eq:decode}
\end{equation}
In a dense multi-class scenario, to obtain the comprehensive multi-class prediction $\mathcal{M}$, the independently generated binary semantic logit maps must first be aggregated. Subsequently, a standard softmax activation is applied to normalize the probabilities across categories, followed by an $\mathtt{argmax}$ operation to derive the final segmentation mask:
\begin{equation}
    \mathcal{M} = \mathrm{Concat}_{i=1}^{N}(\mathcal{M}_{i}),\quad \mathcal{M} \in \mathbb{R}^{N \times h \times w}
.\end{equation}
\paragraph{Limitations}
This pipeline must execute the feature fusion and decoding processes independently for each category, which introduces a inference latency with a complexity of $\mathcal{O}(n)$. Furthermore, the independent prediction paradigm for each category inherently lacks inter-class competition, thereby provoking severe semantic misclassifications and overlapping masks. In contrast, by decoupling the cross-modal fusion process and constructing a dense competition head, our method transforms the native serial binary segmenter into a highly efficient competitive multi-class architecture.

\subsection{Architecture of DCR-SAM}
\label{method:dcr_sam}

\paragraph{Decoupled Visual Conditioning}
To alleviate the prohibitive $\mathcal{O}(n)$ inference latency caused by the concept-specific fusion and decoding process, we design a decoupled visual conditioning strategy. Instead of repeatedly feeding the specific text prompt $p_i$ into the computationally intensive fusion encoder (\cf~\Eq~\ref{eq:fusion}), we introduce a unified learnable prompt sequence $p^{*} \in \mathbb{R}^{l \times d}$. By substituting the concept-specific tokens with $p^{*}$, We completely decouple the deep cross-modal feature fusion from specific prompt tokens. Consequently, the fusion encoder integrates the visual features with the unified prompt sequence through a single forward pass, yielding the joint representation $\mathcal{V}^{*}$:
\begin{equation}
    \mathcal{V}^{*} = E_{fusion}(\mathcal{V}, p^{*}),\quad \mathcal{V}^{*} \in \mathbb{R}^{hw \times d}.
\end{equation}
This modification adapts the native fusion encoder to extract high-quality, globally contextualized visual representations shared across all concepts. These representations serve as a universal visual foundation for all subsequent multi-class prediction tasks. Benefiting from this strictly category-agnostic feature fusion process, our approach fundamentally alleviates the inherent inference latency bottleneck of the native architecture.

\paragraph{Vision-Conditioned Text Refinement}
The native SAM3~\citep{sam3} independent inference paradigm lacks inter-class competition, leading to severe semantic misclassification and overlapping masks among conceptually similar categories. To fundamentally resolve this issue, our DCR-SAM framework shifts from isolated decoding to a dense competitive paradigm, where textual features are explicitly aligned with globally contextualized visual representations to ensure robust cross-modal grounding. However, original textual features extracted by the text encoder are static and context-free, rendering direct alignment suboptimal. Therefore, we propose the \textit{Vision-Conditioned Text Refinement (VCTR)} module, which achieves dynamic alignment by integrating image-specific visual cues into textual features. Specifically, we design a lightweight Cross-Attention (CA) module to refine the text representations. The text tokens $p_{i}$ for each category serve as the \textit{Query} ($Q$), while the globally contextualized visual features $\mathcal{V}^{*}$ act as the \textit{Key} ($K$) and \textit{Value} ($V$). Finally, we aggregate all category features to obtain the aligned textual feature set $\mathcal{\widetilde{P}}=\{\tilde{p}_1, \tilde{p}_2,\dots,\tilde{p}_{N}\}$:
\begin{equation}
    \tilde{p}_{i} = \mathrm{MHA}(\mathrm{LN}(p_i), \mathrm{LN}(\mathcal{V}^*), \mathrm{LN}(\mathcal{V}^*)) + p_i, \quad \mathrm{where}\,\,\tilde{p}_i \in \mathbb{R}^{l \times d},\,\, \mathcal{\widetilde{P}} \in \mathbb{R}^{N \times l \times d}.
    \label{eq:vctr}
\end{equation}
where $\mathrm{MHA}$ and $\mathrm{LN}$ denote the Multi-Head Attention and Layer Normalization, respectively. Notably, this lightweight module involves only simple matrix operations and a minimal number of additional parameters, incurring a negligible inference latency.

\paragraph{Vision-Guided Text Pooling}
For highly efficient dense matching, the aligned text sequence $\mathcal{\widetilde{P}}$ must be compressed into a global feature embedding (\textit{anchor}) for each category. An intuitive approach applies average pooling across the sequence. However, this introduces redundant noise from less relevant content or padding tokens, compromising the discriminative capability of semantic anchors. Instead, we propose the \textit{Vision-Guided Text Pooling (VGTP)} module. First, we apply Global Average Pooling (GAP) and a linear projection layer $\mathcal{F}_{proj}$ to the globally contextualized visual representation $\mathcal{V}^{*}$ to generate a global visual embedding $\mathcal{V}_{g}$:
\begin{equation}
    \mathcal{V}_{g} = \mathcal{F}_{proj}(\mathrm{GAP}(\mathcal{V}^{*})),\quad \mathcal{V}_{g} \in \mathbb{R}^{1 \times d}.
    \label{eq:vgtp_global}
\end{equation}
Next, we compute the attention scores matrix $\mathcal{A}$ between visual embedding and the aligned text sequence. These scores facilitate a weighted aggregation along the sequence length $l$ to generate the final global text anchors $\mathcal{T}_{anchor}$ for each category:
\begin{equation}
    \mathcal{A} = \mathtt{softmax}\left(\frac{\mathcal{\widetilde{P}}\,\mathcal{V}_{g}^{T}}{\sqrt{d}}\right),\quad \mathcal{A} \in \mathbb{R}^{N \times l},
    \label{eq:attention_score}
\end{equation}
\begin{equation}
    \mathcal{T}_{anchor} = \sum_{i=1}^{l} \mathcal{A}_{:, i} \odot \mathcal{\widetilde{P}}_{:, i},\quad \mathcal{T}_{anchor} \in \mathbb{R}^{N \times d}.
    \label{eq:t_anchor}
\end{equation}
\paragraph{Dynamic Background}
In multi-class segmentation tasks, pixels not belonging to predefined target categories are typically assigned to a background class. To adapt the SAM3~\citep{sam3} to this formulation, a straightforward solution uses a generic text prompt, \eg, background. However, this concept is overly broad and lacks explicit visual grounding. Alternatively, hard-coding background categories (\eg, sky or cloud) severely degrades architectural universality, requires extensive dataset-specific engineering, and fails to exhaust all potential unknown distractors. To address these limitations, we introduce an independent, learnable background token $\mathcal{T}_{bg} \in \mathbb{R}^{1 \times d}$ and concatenate it with the generated text anchors to form a complete semantic anchor set $\mathcal{T}$:
\begin{equation}
    \mathcal{T} = \mathrm{Concat}(\mathcal{T}_{anchor}, \mathcal{T}_{bg}),\quad \mathcal{T} \in \mathbb{R}^{(N+1) \times d}.
\end{equation}
\paragraph{Dense Competitive Head}
Equipped with the complete set of text anchors $\mathcal{T}$ and the globally contextualized visual representation $\mathcal{V}^{*}$, we discard the isolated decoding paradigm in favor of a dense competitive head. To ensure stable optimization and prevent feature magnitude discrepancies from dominating the logit computation, we apply $L_2$ normalization along the channel dimension and introduce a temperature scaling parameter $\tau$ prior to the dot product. Specifically, we compute the scaled similarity between the normalized dense visual features and the text anchors to generate the dense logits map $\mathcal{M}^{*}$:
\begin{equation}
    \mathcal{M}^* = \frac{1}{\tau} \left( \frac{\mathcal{T}}{\|\mathcal{T}\|_2} \right) \left( \frac{\mathcal{V}^*}{\|\mathcal{V}^*\|_2} \right)^T,\quad \mathcal{M}^{*} \in \mathbb{R}^{(N+1) \times hw}.
\end{equation}

\subsection{Joint Optimization for DCR-SAM}
\label{method:optimization}

\paragraph{Global Presence Score}
While the dense competitive head suppresses semantic misclassification, relying solely on local dense features can occasionally lead to false positive predictions due to localized visual similarities. To provide a robust constraint, we follow the training strategy of SAM3~\citep{sam3} and introduce a global presence score during the training phase. Specifically, we compute the dot product between the global visual embedding $\mathcal{V}_{g}$ and the set of target semantic anchors $\mathcal{T}_{anchor}$, followed by a $\mathtt{sigmoid}$ operation, to generate the image-level presence scores $S_{presence}$:
\begin{equation}
    S_{presence} = \mathtt{sigmoid}(\mathcal{T}_{anchor}\,\mathcal{V}_{g}^{T}),\quad S_{presence} \in \mathbb{R}^{N \times 1}.
\end{equation}
\paragraph{Joint Objective}
To holistically optimize the proposed model, we formulate a joint end-to-end training objective. We employ the cross-entropy loss and the dice loss to supervise the final mask, and utilize the binary cross-entropy loss to supervise the global presence score, and $\lambda_{i}$ denotes the weight to rescale the loss terms:
\begin{equation}
    \mathcal{L}_{total} = \lambda_{1} \mathcal{L}_{ce} + \lambda_{2} \mathcal{L}_{dice} + \lambda_{3} \mathcal{L}_{bce}^{presence}.
\end{equation}

\section{Experiments}

\subsection{Experimental Setup}

\begin{table}[t]
\caption{\textbf{Comparison with state-of-the-art semantic segmentation methods} on natural scene and remote sensing, with the best results in \textbf{bold}, the second-best \uline{underlined}, and all results evaluated on the validation set. $^\dagger$ denotes \textit{Native-Binary} paradigm and $^\ddagger$ denotes \textit{Native-Multi-Class} paradigm.}
\label{tab:main_res}
\centering
\renewcommand{\arraystretch}{0.8}
\footnotesize
\begin{tabularx}{\textwidth}{l c c c c c c}
\toprule
\multirow{2}{*}{\textbf{Methods}} & \multirow{2}{*}{\textbf{Pub. \& Year}} & \multirow{2}{*}{\textbf{Backbone}} & \multicolumn{4}{c}{\bf \textit{Train on Natural Scene}} \\
\cmidrule(lr){4-7}
& & & \textbf{ADE20K} & \textbf{Cityscapes} & \textbf{COCO-Stuff} & \textbf{Context59} \\
\midrule
DeepLabV3~\cite{deeplabv3}       & ECCV'18    & ResNet-101       & 42.6 & 77.1 & 37.9  & 52.0 \\
DeepLabV3+~\cite{deeplabv3+}     & ECCV'18    & ResNet-101       & 44.5 & 73.6 & --    & 53.2  \\
% SemanticFPN~\cite{semantic_fpn}  & CVPR'19    & ResNet-101       & 40.7 & 77.7 & --    \\
% AlignSeg~\cite{alignseg}         & TPAMI'22   & ResNet-101       & 46.0 & 82.4 & --    \\
HiPoSeg~\cite{HiPoSeg}           & ICLR'26    & ResNet-101       & 49.0 & 84.0 & --    & -- \\
HRNet~\cite{hrnet}               & arXiv'19   & HRNetV2-W48      & 41.9 & 78.5 & 36.0  & 51.9 \\
OCRNet~\cite{ocrnet}             & ECCV'20    & HRNetV2-W48      & 41.5 & 80.0 & 39.0  & 54.4 \\
OCRNet~\cite{ocrnet}             & ECCV'20    & D-ResNet-101     & 45.3 & 81.8 & --    & 54.8 \\
% Mask2Former~\cite{mask2former}   & CVPR'22    & ViT-Adapter-L    & 58.9 & 84.5 & --   \\
% OneFormer~\cite{oneformer}       & CVPR'23    & ConvNeXt-XL      & 57.4 & 83.6 & --   \\
SETR-PUP~\cite{setr}             & CVPR'21    & ViT-L            & 50.1 & 82.2 & --    & 55.3 \\
EoMT~\cite{EoMT}                 & CVPR'25    & ViT-L            & 58.4 & 84.2 & --    & --   \\
% Segmenter~\cite{segmenter}       & ICCV'21    & ViT-L            & 53.6 & 81.3 & --   \\
% Mask2Former~\cite{mask2former}   & CVPR'22    & ViT-L            & 59.4 & 84.8 & --   \\
DTop~\cite{dtop}                 & ICCV'23    & ViT-L            & 52.8 & --   & 47.1  & 62.7 \\
SegViT~\cite{segvit}             & NeurIPS'22 & ViT-L            & 53.3 & --   & 47.4  & 63.0 \\
ECENet~\cite{ecenet}             & ICCV'23    & Swin-L           & 55.2 & 84.5 & --    & 65.9 \\
% Mask2Former~\cite{mask2former}   & CVPR'22    & Swin-L           & 56.1 & 83.3 & --   \\
% Mask DINO~\cite{maskdino}        & CVPR'23    & Swin-L           & 56.6 & 80.0 & --   \\
CPT~\cite{cpt}                   & TIP'25     & Swin-L           & 58.5 & 84.6 & --    & 65.6 \\
SegMAN-L~\cite{segman}           & CVPR'25    & LASS             & 53.2 & 84.2 & --    & --   \\
SegNeXt-T~\cite{segnext}         & NeurIPS'22 & MSCAN-T          & 41.1 & 79.8 & 38.7  & 51.2 \\
% OneFormer~\cite{oneformer}       & CVPR'23    & DiNAT-L          & 58.1 & 83.1 & --   \\
InternImage-H~\cite{internimage} & CVPR'23    & DCNv3            & \uline{60.3} & \uline{86.1} & --    & \uline{70.3} \\
% \hdashline
SAM3$^\dagger$~\cite{sam3}                & ICLR'26    & SAM3 (PE-L+/14) & 46.5 & 75.4 & \uline{48.1} & 66.3 \\
SAM3$^\ddagger$~\cite{sam3}                & ICLR'26    & SAM3 (PE-L+/14) & 41.2 & 72.6 & 45.4  & 64.3 \\
\rowcolor{green!5}
DCR-SAM (Ours)                   & --         & SAM3 (PE-L+/14) & \textbf{68.7\tiny{\textcolor[RGB]{255, 92, 49}{{$\uparrow$}8.4}}} &
                                                                   \textbf{89.1\tiny{\textcolor[RGB]{255, 92, 49}{{$\uparrow$}3.0}}} &
                                                                   \textbf{50.8\tiny{\textcolor[RGB]{255, 92, 49}{{$\uparrow$}2.7}}} &
                                                                   \textbf{72.6\tiny{\textcolor[RGB]{255, 92, 49}{{$\uparrow$}2.3}}} \\
\midrule
\midrule
\multirow{2}{*}{\textbf{Methods}} & \multirow{2}{*}{\textbf{Pub. \& Year}} & \multirow{2}{*}{\textbf{Backbone}} & \multicolumn{4}{c}{\bf \textit{Train on Remote Sensing}} \\
\cmidrule(lr){4-7}
& & & \textbf{LoveDA} & \textbf{Potsdam} & \textbf{Vaihingen} & \textbf{iSAID} \\
\midrule
PSPNet~\cite{pspnet}             & CVPR'17  & ResNet-50        & 48.3 & 82.0 & 76.8 & 60.2 \\
DeepLabV3+~\cite{deeplabv3+}     & ECCV'18  & ResNet-50        & 47.6 & 84.2 & 77.1 & 61.9 \\
DANet~\cite{danet}               & CVPR'19  & ResNet-50        & 50.2 & 81.4 & 77.3 & 60.0 \\
SemanticFPN~\cite{semantic_fpn}  & CVPR'19  & ResNet-50        & 48.2 & 84.6 & 77.9 & 59.3 \\
FarSeg~\cite{farseg}             & CVPR'20  & ResNet-50        & 48.2 & 84.3 & 79.1 & 63.7 \\
LOGCAN++~\cite{logcan++}         & TGRS'25  & ResNet-50        & 53.4 & 87.6 & 83.9 & 65.0 \\
SCSM~\cite{scsm}                 & ISPRS'25 & ResNet-50        & 54.6 & \uline{87.8} & \uline{84.7} & \uline{66.9} \\
Segmenter~\cite{segmenter}       & ICCV'21  & ViT-L            & 47.1 & 86.5 & 79.4 & --   \\
EfficientViT~\cite{efficientvit} & ICCV'23  & EfficientViT-L2  & 47.1 & 84.2 & 80.5 & --   \\
DDP~\cite{ddp}                   & ICCV'23  & Swin-T           & 52.5 & 87.4 & 82.6 & --   \\
% \hdashline
SAM3$^\dagger$~\cite{sam3}                & ICLR'26  & SAM3 (PE-L+/14) & \uline{55.8}          & 66.4            & 69.8          & 61.1 \\
SAM3$^\ddagger$~\cite{sam3}             & ICLR'26  & SAM3 (PE-L+/14) & 50.4          & 61.3            & 65.7          & 55.4 \\
\rowcolor{green!5}
DCR-SAM (Ours)                                                & --       & SAM3 (PE-L+/14) & \textbf{66.5\tiny{\textcolor[RGB]{255, 92, 49}{{$\uparrow$}10.7}}} &
                                                                                              \textbf{90.2\tiny{\textcolor[RGB]{255, 92, 49}{{$\uparrow$}2.4}}} &
                                                                                              \textbf{89.2\tiny{\textcolor[RGB]{255, 92, 49}{{$\uparrow$}4.5}}} & 
                                                                                              \textbf{75.4\tiny{\textcolor[RGB]{255, 92, 49}{{$\uparrow$}8.5}}} \\
\bottomrule
\end{tabularx}
\end{table}

\paragraph{Datasets and Evaluation Metric}
We evaluate our DCR-SAM on four highly challenging natural-scene benchmarks: \textbf{ADE20K}~\citep{ade20k}, which encompasses a dense label space of 150 classes, \textbf{Cityscapes}~\citep{cityscapes}, which contains 19 classes tailored for urban driving scenarios, COCO-Stuff-164K~\citep{cocostuff} (\textbf{COCO-Stuff}), which features a vast and diverse label space of 171 classes, and PASCAL-Context59~\citep{context59} (\textbf{Context59}), which provides dense annotations for 59 prevalent categories. To demonstrate cross-domain robustness, we extend our evaluation to four representative remote sensing benchmarks: \textbf{LoveDA}~\citep{loveda} with 7 classes, ISPRS \textbf{Potsdam}~\citep{potsdam} with 6 classes, ISPRS \textbf{Vaihingen}~\citep{vaihingen} with 6 classes, and \textbf{iSAID}~\citep{isaid} with 16 classes. Unless stated otherwise, all models are trained on their respective training sets, and the results are reported on the official validation sets. We employ the standard mean Intersection over Union (mIoU) as our primary metric.

\paragraph{Baselines}
We compare DCR-SAM with state-of-the-art semantic segmentation methods, including InternImage-H~\citep{internimage}, EoMT~\citep{EoMT}, and SegMAN-L~\citep{segman}. For SAM3~\citep{sam3}, we adopt its complete official architecture, retaining all core components, including the detector and presence prediction modules, and establish two baselines: \ding{192} \textit{Native-Binary} (\textbf{SAM3$^\dagger$}), which follows the official training strategy to independently predict binary masks for each category; and \ding{193} \textit{Native-Multi-Class} (\textbf{SAM3$^\ddagger$}), which concatenates category-wise predictions and is fine-tuned under a standard multi-class training paradigm. Following established SAM3 adaptation practices, semantic masks from both the semantic and detector heads are fused to obtain the final segmentation results. Notably, the latter configuration lacks explicit inter-class competition, potentially leading to semantic overlap among related categories.

\paragraph{Implementation Details}
The proposed DCR-SAM is implemented in the $\mathtt{PyTorch}$ framework~\citep{pytorch}. All models are initialized with the official pre-trained weights of SAM3~\citep{sam3}. For the concept set, we directly utilize the original category names provided by the respective datasets. The length of the unified learnable prompt sequence $p^{*}$ is set to $32$. The scaling parameter $\tau$ is set to 0.07. The scaling weights ($\lambda_1$, $\lambda_2$, and $\lambda_3$) of the loss functions are set to $1$ by default. The dynamic background strategy is optional, it can be omitted when all pixels in the evaluation dataset belong to the target categories. During the training phase, we freeze the original weights of SAM3 backbone and only fine-tune the newly introduced parameters: the learnable prompt sequence, the lightweight cross-modal attention module, and the learnable background token. The network is optimized end-to-end using the AdamW optimizer~\citep{adamw} with an initial learning rate of $1 \times 10^{-4}$ and a weight decay of $5 \times 10^{-4}$. The batch size is set to $8$, and all models are trained for $80,000$ iterations. All experiments are conducted on $8$ NVIDIA RTX 4090 GPUs. Further details are provided in the \Appendix~\ref{appendix:Detailed Implementation Setup}.

\begin{figure}[t]
    \centering
    \includegraphics[width=\linewidth]{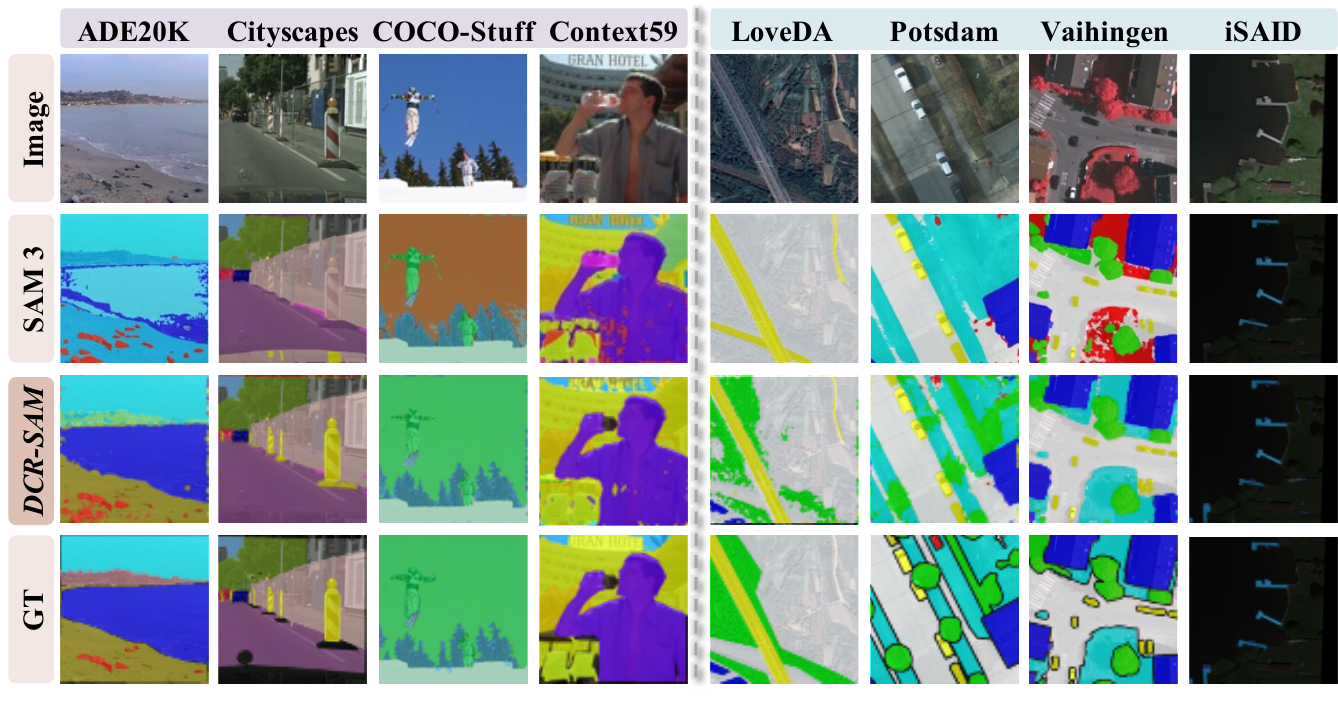}
    \caption{\textbf{Qualitative comparison results} on natural scene and remote sensing. We use the SAM3$^\dagger$ to present the visualization results of the second row.}
    \label{fig:main_vis}
\end{figure}

\subsection{Overall Performance}

\paragraph{Quantitative Comparisons}
As shown in \Table~\ref{tab:main_res}, our method demonstrates exceptional performance across various benchmarks. In natural scene datasets, the proposed method significantly outperforms previous SOTA approaches, yielding substantial improvements ranging from \textbf{2.3\% $\sim$ 8.4\%}. Even compared with InternImage-H~\citep{internimage}, a vision foundation model built upon deformable convolutions, our method consistently achieves superior results. In cross-domain remote sensing scenarios, it continues to demonstrate substantial superiority, achieving improvements of approximately \textbf{2.4\% $\sim$ 10.7\%}. Furthermore, it significantly surpasses specialized methods designed for remote sensing, such as LOGCAN++~\citep{logcan++} and SCSM~\citep{scsm}, verifying its robust scene adaptability. Notably, the performance gains achieved in remote sensing are generally more pronounced than those in natural scenes. We attribute this phenomenon to the significantly smaller concept space of remote sensing imagery compared to natural scenes, enabling the dense competition head to establish more stable semantic anchors. 

Moreover, the comparison with the two SAM3 adaptation baselines demonstrates that both training paradigms fail to achieve optimal results without modifying the core network architecture. Notably, the performance of the \textit{Native-Multi-Class} paradigm (SAM3$^\ddagger$) is inferior to that of the original \textit{Native-Binary} paradigm (SAM3$^\dagger$). This performance degradation indicates that merely adjusting the training strategy fails to compensate for the inherent architectural defects of the model in multi-class segmentation tasks. More experimental analysis in the \Appendix~\ref{appendix:Further Experimental Analysis}.

\paragraph{Qualitative Results}
We visualize the segmentation results across both natural scene and remote sensing benchmarks in \Figure~\ref{fig:main_vis}. SAM3 exhibits limitations such as semantic fragmentation, noisy artifacts (\eg, on the person in Context59), and regional misclassifications (\eg, omitting vegetation in LoveDA) in complex semantic contexts. Conversely, our DCR-SAM preserves the high-fidelity spatial delineation inherent to SAM3 while demonstrating significant improvements in semantic discriminability.

\subsection{Further analysis}

\begin{figure}[t]
    \centering
    
    % ---------------- 左侧：表格 ----------------
    % [c] 表示垂直居中对齐，0.55\textwidth 为表格占用的宽度比例
    \begin{minipage}[c]{0.5\textwidth}
        \centering
        % 使用 \captionof{table} 强制将标题识别为 Table
        \captionof{table}{\textbf{Ablation analysis of core components.} We use the SAM3$^\dagger$ as our baseline model.}
        \label{tab:ablation_study}

        \footnotesize
        \renewcommand{\arraystretch}{0.8}
        \begin{tabularx}{\linewidth}{l Y Y Y Y}
            \toprule
            \multirow{2}{*}{\textbf{Components}} & \multicolumn{2}{c}{\bf ADE20K} & \multicolumn{2}{c}{\bf LoveDA} \\
            \cmidrule(lr){2-3}
            \cmidrule(lr){4-5}
            & \textbf{Time.$\downarrow$} & \textbf{mIoU} & \textbf{Time.$\downarrow$} & \textbf{mIoU} \\
            \midrule
            baseline & 8038 & 46.5 & 647 & 55.8 \\
            \midrule
            \rowcolor{gray!15}
            \multicolumn{5}{l}{\bf \textit{(a) text refinement}} \\ 
            DCR-SAM {\tiny w/o \textit{VCTR}}          & 224  & 61.4 & 179 & 60.9 \\
            \rowcolor{green!5}
            DCR-SAM {\tiny w/ \textit{VCTR}}           & 244  & \textbf{68.7} & 183 & \textbf{66.5} \\
            \midrule
            \rowcolor{gray!15}
            \multicolumn{5}{l}{\bf \textit{(b) pooling strategy}} \\ 
            First        & 244 & 63.1 & 183 & 61.8 \\
            Last         & 244 & 62.8 & 183 & 60.2 \\
            Mean         & 244 & 66.5 & 183 & 63.2 \\
            \rowcolor{green!5}
            \textit{VGTP.}  & 244 & \textbf{68.7} & 183 & \textbf{66.5} \\
            \midrule
            \rowcolor{gray!15}
            \multicolumn{5}{l}{\bf \textit{(c) background strategy}} \\ 
            generic text    & -- & -- & 183 & 59.5 \\
            hard-coding     & -- & -- & 193 & 64.1 \\
            \rowcolor{green!5}
            \textit{dynamic background}       & -- & -- & 183 & \textbf{66.5} \\
            \bottomrule
        \end{tabularx}
    \end{minipage}\hfill % \hfill 用于自动填充中间的间距
    % ---------------- 右侧：图片 ----------------
    % 0.42\textwidth 为图片占用的宽度比例 (0.55 + 0.42 = 0.97，留出3%的中间间隙)
    \begin{minipage}[c]{0.47\textwidth}
        \centering
        \includegraphics[width=0.9\linewidth]{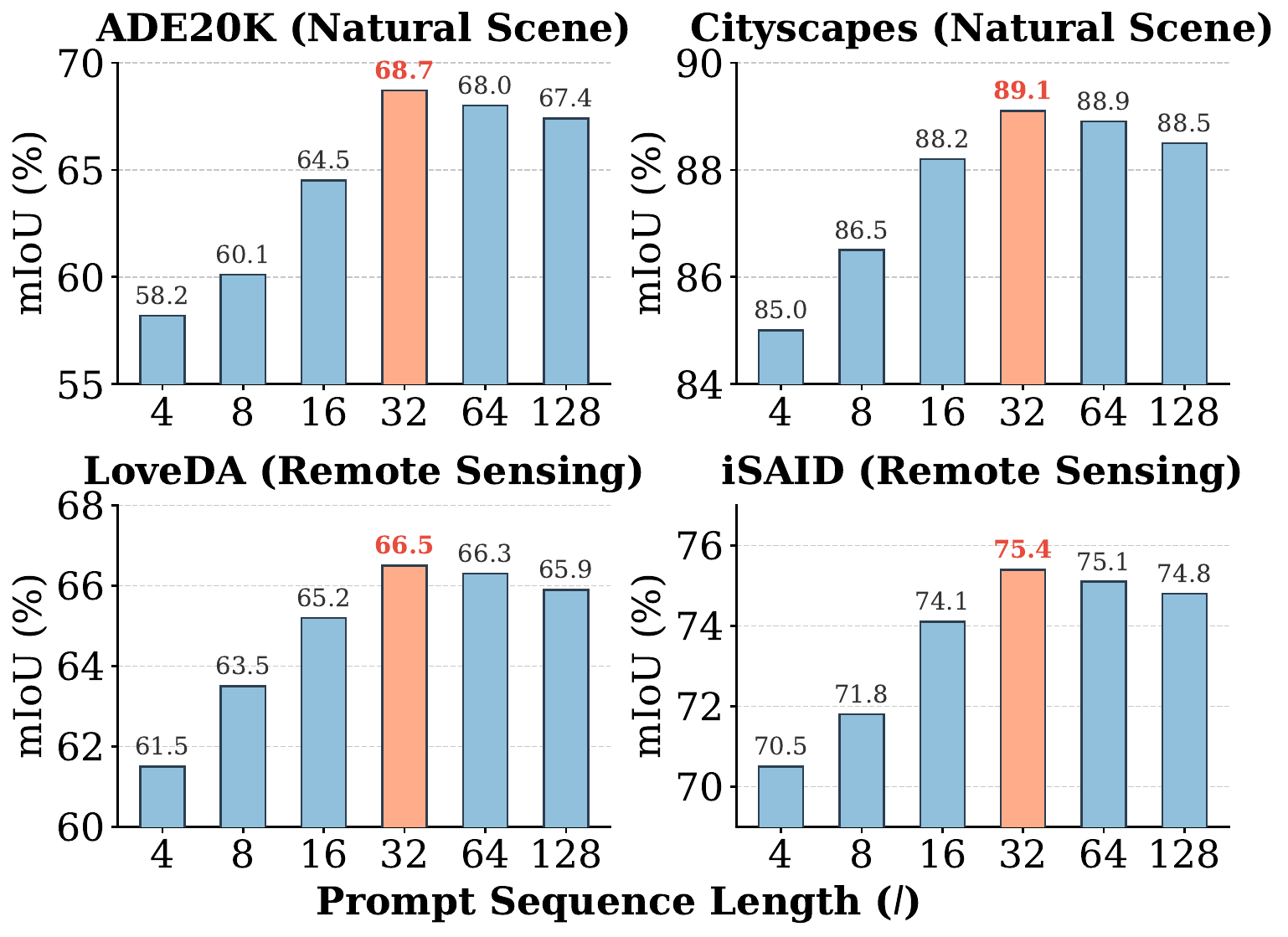}
        \caption{\textbf{Analysis of learnable prompt sequence length.} We select six different lengths ($l\in\{4,8,16,32,64,128\}$) and report the mIoU on four benchmarks. Best results are highlighted in \textcolor[RGB]{255, 172, 138}{orange}.}
        \label{fig:length_albation}
    \end{minipage}
\end{figure}

\paragraph{Component Ablation}
To validate the effectiveness of the proposed modules, we conduct an ablation analysis on the ADE20K and LoveDA benchmarks. \Table~\ref{tab:ablation_study} reports the per-image inference time (ms) and the mIoU on both benchmarks. The proposed method achieves a \textbf{33$\times$} inference speedup on the ADE20K benchmark, decreasing the inference time from \textbf{8038\,ms} to \textbf{244\,ms} alongside substantial performance gains, which demonstrates the necessity of decoupling the cross-modal feature fusion. Furthermore, the text refinement results indicate that incorporating the \textit{VCTR} module yields significant improvements with only a marginal increase in inference latency, extending the inference time from \textbf{224\,ms} to \textbf{244\,ms} on the ADE20K dataset. Regarding the pooling strategy, the proposed \textit{VGTP} module significantly outperforms conventional approaches such as first, last, and mean pooling, confirming that dynamic compression constructs highly discriminative textual anchors. For the background category, it can be observed that the proposed dynamic background strategy ($\mathcal{T}_{bg}$) achieves optimal performance. Although expanding the number of hard-coding background categories can further improve the performance, this strategy requires extensive prompt engineering and manual adaptations specific to the dataset, thereby significantly limiting the general applicability of the model. Additional ablation analyses are detailed in \Appendix~\ref{appendix:Further Experimental Analysis}.

\paragraph{Impact of Learnable Prompt Sequence Length}
We investigate the impact of the learnable prompt sequence length ($l \in \{4, 8, 16, 32, 64, 128\}$) on the final segmentation performance, as illustrated in \Figure~\ref{fig:length_albation}. The performance variations exhibit a highly consistent trend across all four evaluation benchmarks, uniformly achieving optimal results when $l$ is set to 32. When the sequence length is highly restricted ($l \le 8$), the model lacks the representational capacity to capture complex semantic variations, resulting in suboptimal segmentation performance. Conversely, the performance degrades when the sequence length exceeds 32. This degradation is likely attributable to redundant parameters that introduce noise and increase the risk of overfitting. Therefore, we apply a unified sequence length across all benchmarks in this work.

\begin{wrapfigure}{r}{0.42\textwidth}
    \vspace{-13pt}
    \centering
    \setlength{\tabcolsep}{1.3mm}
    \footnotesize
    \renewcommand{\arraystretch}{0.8}
    
    % --- 第一部分：表格 ---
    \captionof{table}{\textbf{Analysis of loss functions.} We report the results on the ADE20K and LoveDA datasets.} % 使用 \captionof 指定为 table
    \label{tab:loss}
    \begin{tabularx}{\linewidth}{c c c c c}
        \toprule
        $\mathcal{L}_{ce}$ & $\mathcal{L}_{dice}$ & $\mathcal{L}_{bce}^{presence}$ & \textbf{ADE20K} & \textbf{LoveDA} \\
        \midrule
        \Checkmark & \XSolidBrush & \XSolidBrush & 66.7 & 64.0 \\
        \Checkmark & \Checkmark & \XSolidBrush & 67.3 & 64.9 \\
        \rowcolor{green!5}
        \Checkmark & \Checkmark & \Checkmark & \textbf{68.7} & \textbf{66.5} \\
        \bottomrule
    \end{tabularx}

    \vspace{1em} % 调整表和图之间的垂直间距，如果想要完全没缝隙可以删掉或设为 0pt

    % --- 第二部分：图片 ---
    \includegraphics[width=0.8\linewidth]{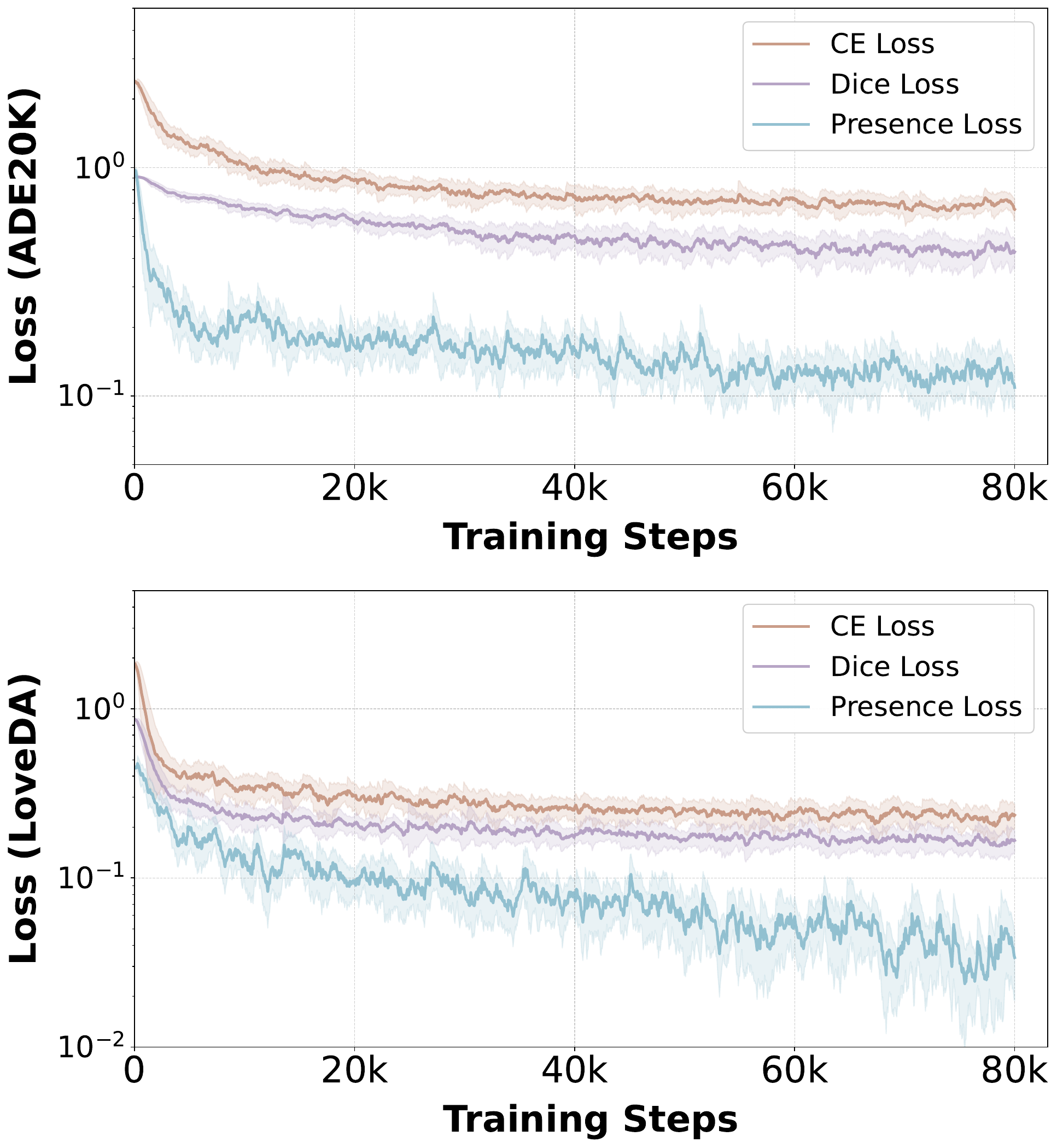}
    \captionof{figure}{\textbf{Loss curves} of three loss functions.}
    \label{fig:loss_curve}
    \vspace{-30pt}
\end{wrapfigure}

\paragraph{Analysis of Objective Function}
We evaluate the individual contributions of the three loss components ($\mathcal{L}_{ce}$, $\mathcal{L}_{dice}$, and $\mathcal{L}_{bce}^{presence}$) in \Table~\ref{tab:loss}. The results indicate that $\mathcal{L}_{ce}$ and $\mathcal{L}_{dice}$ are highly complementary, and their joint application yields substantial improvements. Ultimately, the incorporation of the binary cross-entropy loss ($\mathcal{L}_{bce}^{presence}$) for the prediction of category presence achieves peak mIoU scores of 68.7\% and 66.5\% on the two benchmarks, respectively. This demonstrates the critical role of this component in mitigating false positives. In addition, \Figure~\ref{fig:loss_curve} visualizes the training trajectories, which illustrate a highly stable and synchronous convergence process across both domains, thereby validating the robustness of the joint optimization strategy.

\paragraph{Comparison with SAM 3-Based Methods.}
We further compare DCR-SAM with four recent SAM 3-based methods, including SegEarth-OV3~\citep{segearthov3}, CoCo-SAM3~\citep{coco-sam3}, EfficientSAM3~\citep{efficientsam3}, and SAM3-Adapter~\citep{sam3-adapter}. As shown in \Table~\ref{tab:sam3_comparison}, DCR-SAM consistently outperforms these methods, achieving 68.7\% mIoU on ADE20K and 66.5\% mIoU on LoveDA, surpassing the strongest baselines by 21.5\% and 13.6\%, respectively. While SegEarth-OV3 and CoCo-SAM3 adopt training-free strategies, EfficientSAM3 and SAM3-Adapter primarily focus on efficient adaptation without explicitly addressing inter-class semantic conflicts. In contrast, DCR-SAM introduces a dense competitive head to enforce explicit inter-class competition, effectively mitigating semantic overlap and improving multi-class segmentation performance.

\begin{table}[t]
    \centering
    \footnotesize

    % Left: Comparison with SAM 3-Based Methods
    \begin{minipage}[t]{0.48\linewidth}
        \vspace{0pt}
        \centering
        \begin{minipage}[t][4\baselineskip][t]{\linewidth}
            \caption{\textbf{Comparison with SAM 3-based methods} on ADE20K and LoveDA (mIoU, \%).}
            \label{tab:sam3_comparison}
        \end{minipage}
        \par\vspace{3pt}
        \begin{tabularx}{\linewidth}{@{}lYY@{}}
        \toprule
        \textbf{Method} & \textbf{ADE20K} & \textbf{LoveDA} \\
        \midrule
        SegEarth-OV3 & 36.6 & 47.4 \\
        CoCo-SAM3 & 38.3 & -- \\
        EfficientSAM3 & 43.2 & 50.3 \\
        SAM3-Adapter & 47.2 & 52.9 \\
        \midrule
        \rowcolor{green!5}
        \textbf{\textit{DCR-SAM}} & \textbf{68.7} & \textbf{66.5} \\
        \bottomrule
        \end{tabularx}
    \end{minipage}
    \hfill
    % Right: Comparison under a Unified Backbone
    \begin{minipage}[t]{0.48\linewidth}
        \vspace{0pt}
        \centering
        \begin{minipage}[t][4\baselineskip][t]{\linewidth}
            \caption{\textbf{Comparison} of different segmentation architectures using a unified PE-L+/14 backbone on ADE20K (mIoU, \%).}
            \label{tab:unified_backbone}
        \end{minipage}
        \par\vspace{3pt}
        \begingroup
        \renewcommand{\arraystretch}{1.20}
        \begin{tabularx}{\linewidth}{@{}lYY@{}}
        \toprule
        \textbf{Method} & \textbf{Backbone} & \textbf{mIoU$\uparrow$} \\
        \midrule
        UPerNet & PE-L+/14 & 59.6 \\
        EoMT-Style & PE-L+/14 & 52.1 \\
        Mask2Former & PE-L+/14 & 61.3 \\
        \midrule
        \rowcolor{green!5}
        \textbf{\textit{DCR-SAM}} & \textbf{PE-L+/14} & \textbf{68.7} \\
        \bottomrule
        \end{tabularx}
        \endgroup
    \end{minipage}
\end{table}

\paragraph{Comparison under a Unified Backbone}
To isolate the contribution of our segmentation architecture, we further compare DCR-SAM with representative segmentation heads, including UPerNet~\citep{upernet}, Mask2Former~\citep{mask2former}, and EoMT-Style~\citep{EoMT}, using the same PE-L+/14 backbone~\citep{pe}. As shown in \Table~\ref{tab:unified_backbone}, DCR-SAM achieves 68.7 mIoU on ADE20K, outperforming UPerNet, EoMT-Style, and Mask2Former by 9.1\%, 16.6\%, and 7.4\%, respectively. These results demonstrate that the performance gains are not solely attributable to the backbone, but also benefit from our proposed dense competitive architecture. Notably, the PE backbone is specifically optimized for open-vocabulary tasks, making it less suitable for conventional closed-set segmentation heads. By preserving the native promptable architecture of SAM 3 and introducing explicit inter-class competition, DCR-SAM better exploits its pre-trained cross-modal representations for dense multi-class segmentation.

\section{Conclusion}
In this paper, we propose the \textbf{D}ense \textbf{C}ompetitive \textbf{R}epresentations \textbf{SAM} (\textbf{DCR-SAM}) framework to mitigate the inference latency and inherent semantic misclassifications encountered during the adaptation of SAM3 to multi-class segmentation. The proposed method decouples the feature fusion process to ensure optimal inference efficiency. Furthermore, the alignment of textual and visual representations within a dense competitive paradigm effectively suppresses overlapping predictions among conceptually similar categories. Extensive experiments demonstrate that the framework achieves SOTA performance across diverse benchmarks while delivering a 33$\times$ speedup. We hope our work paves the way for adapting foundation models to complex, real-world multi-class segmentation tasks.

\paragraph{Limitations} DCR-SAM reduces text prompt sensitivity and inference latency, facilitating its potential application in autonomous systems and medical imaging. However, its reliance on large-scale foundation models incurs substantial computational costs and may inherit biases from pre-training data. Moreover, inter-class competition may impair the recognition of unseen concepts. Future work will focus on improving computational efficiency and generalization in open-vocabulary segmentation.

% \section*{Acknowledgment}

\bibliographystyle{plainnat}
\bibliography{neurips2026}

@inproceedings{pspnet,
  title={Pyramid scene parsing network},
  author={Zhao, Hengshuang and Shi, Jianping and Qi, Xiaojuan and Wang, Xiaogang and Jia, Jiaya},
  booktitle={Proceedings of the IEEE conference on computer vision and pattern recognition},
  pages={2881--2890},
  year={2017}
}

@inproceedings{deeplabv3+,
  title={Encoder-decoder with atrous separable convolution for semantic image segmentation},
  author={Chen, Liang-Chieh and Zhu, Yukun and Papandreou, George and Schroff, Florian and Adam, Hartwig},
  booktitle={Proceedings of the European conference on computer vision (ECCV)},
  pages={801--818},
  year={2018}
}

@inproceedings{danet,
  title={Dual attention network for scene segmentation},
  author={Fu, Jun and Liu, Jing and Tian, Haijie and Li, Yong and Bao, Yongjun and Fang, Zhiwei and Lu, Hanqing},
  booktitle={Proceedings of the IEEE/CVF conference on computer vision and pattern recognition},
  pages={3146--3154},
  year={2019}
}

@inproceedings{semantic_fpn,
  title={Panoptic feature pyramid networks},
  author={Kirillov, Alexander and Girshick, Ross and He, Kaiming and Doll{\'a}r, Piotr},
  booktitle={Proceedings of the IEEE/CVF conference on computer vision and pattern recognition},
  pages={6399--6408},
  year={2019}
}

@inproceedings{farseg,
  title={Foreground-aware relation network for geospatial object segmentation in high spatial resolution remote sensing imagery},
  author={Zheng, Zhuo and Zhong, Yanfei and Wang, Junjue and Ma, Ailong},
  booktitle={Proceedings of the IEEE/CVF conference on computer vision and pattern recognition},
  pages={4096--4105},
  year={2020}
}

@inproceedings{efficientvit,
  title={Efficientvit: Lightweight multi-scale attention for high-resolution dense prediction},
  author={Cai, Han and Li, Junyan and Hu, Muyan and Gan, Chuang and Han, Song},
  booktitle={Proceedings of the IEEE/CVF international conference on computer vision},
  pages={17302--17313},
  year={2023}
}

@inproceedings{ddp,
  title={Ddp: Diffusion model for dense visual prediction},
  author={Ji, Yuanfeng and Chen, Zhe and Xie, Enze and Hong, Lanqing and Liu, Xihui and Liu, Zhaoqiang and Lu, Tong and Li, Zhenguo and Luo, Ping},
  booktitle={Proceedings of the IEEE/CVF International Conference on Computer Vision},
  pages={21741--21752},
  year={2023}
}

@article{logcan++,
  title={Logcan++: Adaptive local-global class-aware network for semantic segmentation of remote sensing images},
  author={Ma, Xiaowen and Lian, Rongrong and Wu, Zhenkai and Guo, Hongbo and Yang, Fan and Ma, Mengting and Wu, Sensen and Du, Zhenhong and Zhang, Wei and Song, Siyang},
  journal={IEEE Transactions on Geoscience and Remote Sensing},
  year={2025},
  publisher={IEEE}
}

@article{scsm,
  title={A novel scene coupling semantic mask network for remote sensing image segmentation},
  author={Ma, Xiaowen and Lian, Rongrong and Wu, Zhenkai and Guan, Renxiang and Hong, Tingfeng and Zhao, Mengjiao and Ma, Mengting and Nie, Jiangtao and Du, Zhenhong and Song, Siyang and others},
  journal={ISPRS Journal of Photogrammetry and Remote Sensing},
  volume={221},
  pages={44--63},
  year={2025},
  publisher={Elsevier}
}

@article{sam3,
  title={Sam 3: Segment anything with concepts},
  author={Carion, Nicolas and Gustafson, Laura and Hu, Yuan-Ting and Debnath, Shoubhik and Hu, Ronghang and Suris, Didac and Ryali, Chaitanya and Alwala, Kalyan Vasudev and Khedr, Haitham and Huang, Andrew and others},
  journal={arXiv preprint arXiv:2511.16719},
  year={2025}
}

@inproceedings{mask2former,
  title={Masked-attention mask transformer for universal image segmentation},
  author={Cheng, Bowen and Misra, Ishan and Schwing, Alexander G and Kirillov, Alexander and Girdhar, Rohit},
  booktitle={Proceedings of the IEEE/CVF conference on computer vision and pattern recognition},
  pages={1290--1299},
  year={2022}
}

@inproceedings{dtop,
  title={Dynamic token pruning in plain vision transformers for semantic segmentation},
  author={Tang, Quan and Zhang, Bowen and Liu, Jiajun and Liu, Fagui and Liu, Yifan},
  booktitle={Proceedings of the IEEE/CVF International Conference on Computer Vision},
  pages={777--786},
  year={2023}
}

@inproceedings{internimage,
  title={Internimage: Exploring large-scale vision foundation models with deformable convolutions},
  author={Wang, Wenhai and Dai, Jifeng and Chen, Zhe and Huang, Zhenhang and Li, Zhiqi and Zhu, Xizhou and Hu, Xiaowei and Lu, Tong and Lu, Lewei and Li, Hongsheng and others},
  booktitle={Proceedings of the IEEE/CVF conference on computer vision and pattern recognition},
  pages={14408--14419},
  year={2023}
}

@inproceedings{segman,
  title={SegMAN: Omni-scale context modeling with state space models and local attention for semantic segmentation},
  author={Fu, Yunxiang and Lou, Meng and Yu, Yizhou},
  booktitle={Proceedings of the Computer Vision and Pattern Recognition Conference},
  pages={19077--19087},
  year={2025}
}

@inproceedings{EoMT,
  title={Your vit is secretly an image segmentation model},
  author={Kerssies, Tommie and Cavagnero, Niccolo and Hermans, Alexander and Norouzi, Narges and Averta, Giuseppe and Leibe, Bastian and Dubbelman, Gijs and De Geus, Daan},
  booktitle={Proceedings of the computer vision and pattern recognition conference},
  pages={25303--25313},
  year={2025}
}

@inproceedings{HiPoSeg,
  title={Hierarchical Prototype Learning for Semantic Segmentation},
  author={Lim, Seoha and Kim, Jinmyeong and Kim, Jieun and Cho, Sung-Bae},
  booktitle={The Fourteenth International Conference on Learning Representations},
  year={2025}
}

@article{cpt,
  title={Rethinking feature reconstruction via category prototype in semantic segmentation},
  author={Tang, Quan and Liu, Chuanjian and Liu, Fagui and Jiang, Jun and Zhang, Bowen and Chen, CL Philip and Han, Kai and Wang, Yunhe},
  journal={IEEE Transactions on Image Processing},
  volume={34},
  pages={1036--1047},
  year={2025},
  publisher={IEEE}
}

@inproceedings{ocrnet,
  title={Object-contextual representations for semantic segmentation},
  author={Yuan, Yuhui and Chen, Xilin and Wang, Jingdong},
  booktitle={European conference on computer vision},
  pages={173--190},
  year={2020},
  organization={Springer}
}

@inproceedings{setr,
  title={Rethinking semantic segmentation from a sequence-to-sequence perspective with transformers},
  author={Zheng, Sixiao and Lu, Jiachen and Zhao, Hengshuang and Zhu, Xiatian and Luo, Zekun and Wang, Yabiao and Fu, Yanwei and Feng, Jianfeng and Xiang, Tao and Torr, Philip HS and others},
  booktitle={Proceedings of the IEEE/CVF conference on computer vision and pattern recognition},
  pages={6881--6890},
  year={2021}
}

@inproceedings{segmenter,
  title={Segmenter: Transformer for semantic segmentation},
  author={Strudel, Robin and Garcia, Ricardo and Laptev, Ivan and Schmid, Cordelia},
  booktitle={Proceedings of the IEEE/CVF international conference on computer vision},
  pages={7262--7272},
  year={2021}
}

@inproceedings{ecenet,
  title={Boosting semantic segmentation from the perspective of explicit class embeddings},
  author={Liu, Yuhe and Liu, Chuanjian and Han, Kai and Tang, Quan and Qin, Zengchang},
  booktitle={Proceedings of the IEEE/CVF International Conference on Computer Vision},
  pages={821--831},
  year={2023}
}

@inproceedings{ade20k,
  title={Scene parsing through ade20k dataset},
  author={Zhou, Bolei and Zhao, Hang and Puig, Xavier and Fidler, Sanja and Barriuso, Adela and Torralba, Antonio},
  booktitle={Proceedings of the IEEE conference on computer vision and pattern recognition},
  pages={633--641},
  year={2017}
}

@inproceedings{cityscapes,
  title={The cityscapes dataset for semantic urban scene understanding},
  author={Cordts, Marius and Omran, Mohamed and Ramos, Sebastian and Rehfeld, Timo and Enzweiler, Markus and Benenson, Rodrigo and Franke, Uwe and Roth, Stefan and Schiele, Bernt},
  booktitle={Proceedings of the IEEE conference on computer vision and pattern recognition},
  pages={3213--3223},
  year={2016}
}

@article{loveda,
  title={LoveDA: A remote sensing land-cover dataset for domain adaptive semantic segmentation},
  author={Wang, Junjue and Zheng, Zhuo and Ma, Ailong and Lu, Xiaoyan and Zhong, Yanfei},
  journal={arXiv preprint arXiv:2110.08733},
  year={2021}
}

@article{potsdam,
  title={ISPRS semantic labeling contest},
  author={Rottensteiner, Franz and Sohn, Gunho and Gerke, Markus and Wegner, Jan D},
  journal={ISPRS: Leopoldsh{\"o}he, Germany},
  volume={1},
  number={4},
  pages={4},
  year={2014}
}

@misc{vaihingen,
  title={Use of the stair vision library within the ISPRS 2D semantic labeling benchmark (Vaihingen)},
  author={Markus Gerke, ITC},
  year={2014},
  publisher={University of Twente Enschede, The Netherlands}
}

@inproceedings{isaid,
  title={isaid: A large-scale dataset for instance segmentation in aerial images},
  author={Waqas Zamir, Syed and Arora, Aditya and Gupta, Akshita and Khan, Salman and Sun, Guolei and Shahbaz Khan, Fahad and Zhu, Fan and Shao, Ling and Xia, Gui-Song and Bai, Xiang},
  booktitle={Proceedings of the IEEE/CVF conference on computer vision and pattern recognition workshops},
  pages={28--37},
  year={2019}
}

@article{deeplabv3,
  title={Rethinking atrous convolution for semantic image segmentation},
  author={Chen, Liang-Chieh and Papandreou, George and Schroff, Florian and Adam, Hartwig},
  journal={arXiv preprint arXiv:1706.05587},
  year={2017}
}

@article{hrnet,
  title={High-resolution representations for labeling pixels and regions},
  author={Sun, Ke and Zhao, Yang and Jiang, Borui and Cheng, Tianheng and Xiao, Bin and Liu, Dong and Mu, Yadong and Wang, Xinggang and Liu, Wenyu and Wang, Jingdong},
  journal={arXiv preprint arXiv:1904.04514},
  year={2019}
}

@article{segvit,
  title={Segvit: Semantic segmentation with plain vision transformers},
  author={Zhang, Bowen and Tian, Zhi and Tang, Quan and Chu, Xiangxiang and Wei, Xiaolin and Shen, Chunhua and others},
  journal={Advances in Neural Information Processing Systems},
  volume={35},
  pages={4971--4982},
  year={2022}
}

@article{segnext,
  title={Segnext: Rethinking convolutional attention design for semantic segmentation},
  author={Guo, Meng-Hao and Lu, Cheng-Ze and Hou, Qibin and Liu, Zhengning and Cheng, Ming-Ming and Hu, Shi-Min},
  journal={Advances in neural information processing systems},
  volume={35},
  pages={1140--1156},
  year={2022}
}

@inproceedings{cocostuff,
  title={Coco-stuff: Thing and stuff classes in context},
  author={Caesar, Holger and Uijlings, Jasper and Ferrari, Vittorio},
  booktitle={Proceedings of the IEEE conference on computer vision and pattern recognition},
  pages={1209--1218},
  year={2018}
}

@inproceedings{context59,
  title={The role of context for object detection and semantic segmentation in the wild},
  author={Mottaghi, Roozbeh and Chen, Xianjie and Liu, Xiaobai and Cho, Nam-Gyu and Lee, Seong-Whan and Fidler, Sanja and Urtasun, Raquel and Yuille, Alan},
  booktitle={Proceedings of the IEEE conference on computer vision and pattern recognition},
  pages={891--898},
  year={2014}
}

@article{pytorch,
  title={Pytorch: An imperative style, high-performance deep learning library},
  author={Paszke, Adam and Gross, Sam and Massa, Francisco and Lerer, Adam and Bradbury, James and Chanan, Gregory and Killeen, Trevor and Lin, Zeming and Gimelshein, Natalia and Antiga, Luca and others},
  journal={Advances in neural information processing systems},
  volume={32},
  year={2019}
}

@article{adamw,
  title={Decoupled weight decay regularization},
  author={Loshchilov, Ilya and Hutter, Frank},
  journal={arXiv preprint arXiv:1711.05101},
  year={2017}
}

@inproceedings{sam,
  title={Segment anything},
  author={Kirillov, Alexander and Mintun, Eric and Ravi, Nikhila and Mao, Hanzi and Rolland, Chloe and Gustafson, Laura and Xiao, Tete and Whitehead, Spencer and Berg, Alexander C and Lo, Wan-Yen and others},
  booktitle={Proceedings of the IEEE/CVF international conference on computer vision},
  pages={4015--4026},
  year={2023}
}

@article{sam2,
  title={Sam 2: Segment anything in images and videos},
  author={Ravi, Nikhila and Gabeur, Valentin and Hu, Yuan-Ting and Hu, Ronghang and Ryali, Chaitanya and Ma, Tengyu and Khedr, Haitham and R{\"a}dle, Roman and Rolland, Chloe and Gustafson, Laura and others},
  journal={arXiv preprint arXiv:2408.00714},
  year={2024}
}

@article{segearthov3,
  title={SegEarth-OV3: Exploring SAM 3 for Open-Vocabulary Semantic Segmentation in Remote Sensing Images},
  author={Li, Kaiyu and Zhang, Shengqi and Deng, Yupeng and Wang, Zhi and Meng, Deyu and Cao, Xiangyong},
  journal={arXiv preprint arXiv:2512.08730},
  year={2025}
}

@article{blushtein2025effectiveness,
  title={On the Effectiveness of Textual Prompting with Lightweight Fine-Tuning for SAM3 Remote Sensing Segmentation},
  author={Blushtein-Livnon, Roni and Rafaeli, Osher and Ioffe, David and Boger, Amir and Esquenazi, Karen Sandberg and Svoray, Tal},
  journal={arXiv preprint arXiv:2512.15564},
  year={2025}
}

@article{tang2026cad,
  title={CAD-Prompted SAM3: Geometry-Conditioned Instance Segmentation for Industrial Objects},
  author={Tang, Zhenran and Nagabhirava, Rohan and Liu, Changliu},
  journal={arXiv preprint arXiv:2602.20551},
  year={2026}
}

@article{niehorster2026eye,
  title={Eye image segmentation using visual and concept prompts with Segment Anything Model 3 (SAM3)},
  author={Niehorster, Diederick C and Nystr{\"o}m, Marcus},
  journal={arXiv preprint arXiv:2603.17715},
  year={2026}
}

@inproceedings{bundele2025memory,
  title={Memory-Enhanced SAM3 for Occlusion-Robust Surgical Instrument Segmentation},
  author={Bundele, Valay and Hosseinzadeh, Mehran and Lensch, Hendrik},
  booktitle={Medical Imaging with Deep Learning},
  year={2025}
}

@article{lin2026progressive,
  title={Progressive Per-Branch Depth Optimization for DEFOM-Stereo and SAM3 Joint Analysis in UAV Forestry Applications},
  author={Lin, Yida and Xue, Bing and Zhang, Mengjie and Schofield, Sam and Green, Richard},
  journal={arXiv preprint arXiv:2602.20539},
  year={2026}
}

@article{sam3-litetext,
  title={SAM3-LiteText: An Anatomical Study of the SAM3 Text Encoder for Efficient Vision-Language Segmentation},
  author={Zeng, Chengxi and Jiang, Yuxuan and Gao, Ge and Wang, Shuai and Danier, Duolikun and Zhu, Bin and Rudinac, Stevan and Bull, David and Zhang, Fan},
  journal={arXiv preprint arXiv:2602.12173},
  year={2026}
}

@article{sam3-i,
  title={SAM3-I: Segment Anything with Instructions},
  author={Li, Jingjing and Feng, Yue and Guo, Yuchen and Huang, Jincai and Piao, Yongri and Bi, Qi and Zhang, Miao and Zhao, Xiaoqi and Chen, Qiang and Zou, Shihao and others},
  journal={arXiv preprint arXiv:2512.04585},
  year={2025}
}

@article{concept-bank,
  title={Taming SAM3 in the Wild: A Concept Bank for Open-Vocabulary Segmentation},
  author={Pei, Gensheng and Jiang, Xiruo and Yao, Yazhou and Shu, Xiangbo and Shen, Fumin and Jeon, Byeungwoo},
  journal={arXiv preprint arXiv:2602.06333},
  year={2026}
}

@inproceedings{fcn,
  title={Fully convolutional networks for semantic segmentation},
  author={Long, Jonathan and Shelhamer, Evan and Darrell, Trevor},
  booktitle={Proceedings of the IEEE conference on computer vision and pattern recognition},
  pages={3431--3440},
  year={2015}
}

@inproceedings{unet,
  title={U-net: Convolutional networks for biomedical image segmentation},
  author={Ronneberger, Olaf and Fischer, Philipp and Brox, Thomas},
  booktitle={International Conference on Medical image computing and computer-assisted intervention},
  pages={234--241},
  year={2015},
  organization={Springer}
}

@article{segformer,
  title={SegFormer: Simple and efficient design for semantic segmentation with transformers},
  author={Xie, Enze and Wang, Wenhai and Yu, Zhiding and Anandkumar, Anima and Alvarez, Jose M and Luo, Ping},
  journal={Advances in neural information processing systems},
  volume={34},
  pages={12077--12090},
  year={2021}
}

@inproceedings{clip,
  title={Learning transferable visual models from natural language supervision},
  author={Radford, Alec and Kim, Jong Wook and Hallacy, Chris and Ramesh, Aditya and Goh, Gabriel and Agarwal, Sandhini and Sastry, Girish and Askell, Amanda and Mishkin, Pamela and Clark, Jack and others},
  booktitle={International conference on machine learning},
  pages={8748--8763},
  year={2021},
  organization={PmLR}
}

@inproceedings{siglip,
  title={Sigmoid loss for language image pre-training},
  author={Zhai, Xiaohua and Mustafa, Basil and Kolesnikov, Alexander and Beyer, Lucas},
  booktitle={Proceedings of the IEEE/CVF international conference on computer vision},
  pages={11975--11986},
  year={2023}
}

@inproceedings{align,
  title={Scaling up visual and vision-language representation learning with noisy text supervision},
  author={Jia, Chao and Yang, Yinfei and Xia, Ye and Chen, Yi-Ting and Parekh, Zarana and Pham, Hieu and Le, Quoc and Sung, Yun-Hsuan and Li, Zhen and Duerig, Tom},
  booktitle={International conference on machine learning},
  pages={4904--4916},
  year={2021},
  organization={PMLR}
}

@inproceedings{dino,
  title={Emerging properties in self-supervised vision transformers},
  author={Caron, Mathilde and Touvron, Hugo and Misra, Ishan and J{\'e}gou, Herv{\'e} and Mairal, Julien and Bojanowski, Piotr and Joulin, Armand},
  booktitle={Proceedings of the IEEE/CVF international conference on computer vision},
  pages={9650--9660},
  year={2021}
}

@article{dinov2,
  title={Dinov2: Learning robust visual features without supervision},
  author={Oquab, Maxime and Darcet, Timoth{\'e}e and Moutakanni, Th{\'e}o and Vo, Huy and Szafraniec, Marc and Khalidov, Vasil and Fernandez, Pierre and Haziza, Daniel and Massa, Francisco and El-Nouby, Alaaeldin and others},
  journal={arXiv preprint arXiv:2304.07193},
  year={2023}
}

@article{dinov3,
  title={Dinov3},
  author={Sim{\'e}oni, Oriane and Vo, Huy V and Seitzer, Maximilian and Baldassarre, Federico and Oquab, Maxime and Jose, Cijo and Khalidov, Vasil and Szafraniec, Marc and Yi, Seungeun and Ramamonjisoa, Micha{\"e}l and others},
  journal={arXiv preprint arXiv:2508.10104},
  year={2025}
}

@article{maskformer,
  title={Per-pixel classification is not all you need for semantic segmentation},
  author={Cheng, Bowen and Schwing, Alex and Kirillov, Alexander},
  journal={Advances in neural information processing systems},
  volume={34},
  pages={17864--17875},
  year={2021}
}

@article{sam3-adapter,
  title={SAM3-Adapter: Efficient Adaptation of Segment Anything 3 for Camouflage Object Segmentation, Shadow Detection, and Medical Image Segmentation},
  author={Chen, Tianrun and Cao, Runlong and Yu, Xinda and Zhu, Lanyun and Ding, Chaotao and Ji, Deyi and Chen, Cheng and Zhu, Qi and Xu, Chunyan and Mao, Papa and others},
  journal={arXiv preprint arXiv:2511.19425},
  year={2025}
}

@article{efficientsam3,
  title={EfficientSAM3: Progressive Hierarchical Distillation for Video Concept Segmentation from SAM1, 2, and 3},
  author={Zeng, Chengxi and Jiang, Yuxuan and Zhang, Aaron},
  journal={arXiv preprint arXiv:2511.15833},
  year={2025}
}

@article{coco-sam3,
  title={CoCo-SAM3: Harnessing Concept Conflict in Open-Vocabulary Semantic Segmentation},
  author={Chen, Yanhui and Yang, Baoyao and Liu, Siqi and Wang, Jingchao},
  journal={arXiv preprint arXiv:2604.19648},
  year={2026}
}

@article{densecrf,
  title={Efficient inference in fully connected crfs with gaussian edge potentials},
  author={Kr{\"a}henb{\"u}hl, Philipp and Koltun, Vladlen},
  journal={Advances in neural information processing systems},
  volume={24},
  year={2011}
}

@inproceedings{pamr,
  title={Single-stage semantic segmentation from image labels},
  author={Araslanov, Nikita and Roth, Stefan},
  booktitle={Proceedings of the IEEE/CVF conference on computer vision and pattern recognition},
  pages={4253--4262},
  year={2020}
}

@article{gpt-5,
  title={Early science acceleration experiments with GPT-5},
  author={Bubeck, S{\'e}bastien and Coester, Christian and Eldan, Ronen and Gowers, Timothy and Lee, Yin Tat and Lupsasca, Alexandru and Sawhney, Mehtaab and Scherrer, Robert and Sellke, Mark and Spears, Brian K and others},
  journal={arXiv preprint arXiv:2511.16072},
  year={2025}
}

@article{test-time-contrast,
  title={Test-time Contrastive Concepts for Open-world Semantic Segmentation with Vision-Language Models},
  author={Wysocza{\'n}ska, Monika and Vobecky, Antonin and Cardiel, Amaia and Trzci{\'n}ski, Tomasz and Marlet, Renaud and Bursuc, Andrei and Sim{\'e}oni, Oriane},
  journal={arXiv preprint arXiv:2407.05061},
  year={2024}
}

@article{layernorm,
  title={Layer normalization},
  author={Ba, Jimmy Lei and Kiros, Jamie Ryan and Hinton, Geoffrey E},
  journal={arXiv preprint arXiv:1607.06450},
  year={2016}
}

@inproceedings{resnet,
  title={Deep residual learning for image recognition},
  author={He, Kaiming and Zhang, Xiangyu and Ren, Shaoqing and Sun, Jian},
  booktitle={Proceedings of the IEEE conference on computer vision and pattern recognition},
  pages={770--778},
  year={2016}
}

@incollection{bilinear,
  title={Bilinear interpolation},
  author={Kirkland, Earl J},
  booktitle={Advanced computing in electron microscopy},
  pages={261--263},
  year={2010},
  publisher={Springer}
}

@inproceedings{spanet,
  title={Spanet: Spatial pyramid attention network for enhanced image recognition},
  author={Guo, Jingda and Ma, Xu and Sansom, Andrew and McGuire, Mara and Kalaani, Andrew and Chen, Qi and Tang, Sihai and Yang, Qing and Fu, Song},
  booktitle={2020 IEEE International Conference on Multimedia and Expo (ICME)},
  year={2020}
}

@article{attention,
  title={Attention is all you need},
  author={Vaswani, Ashish and Shazeer, Noam and Parmar, Niki and Uszkoreit, Jakob and Jones, Llion and Gomez, Aidan N and Kaiser, {\L}ukasz and Polosukhin, Illia},
  journal={Advances in neural information processing systems},
  volume={30},
  year={2017}
}

@inproceedings{upernet,
  title={Unified perceptual parsing for scene understanding},
  author={Xiao, Tete and Liu, Yingcheng and Zhou, Bolei and Jiang, Yuning and Sun, Jian},
  booktitle={European conference on computer vision},
  pages={432--448},
  year={2018},
  organization={Springer}
}

@article{pe,
  title={Perception encoder: The best visual embeddings are not at the output of the network},
  author={Bolya, Daniel and Huang, Po-Yao and Sun, Peize and Cho, Jang Hyun and Madotto, Andrea and Wei, Chen and Ma, Tengyu and Zhi, Jiale and Rajasegaran, Jathushan and Bangalath, Hanoona and others},
  journal={Advances in Neural Information Processing Systems},
  volume={38},
  pages={60884--60937},
  year={2026}
}

%%%%%%%%%%%%%%%%%%%%%%%%%%%%%%%%%%%%%%%%%%%%%%%%%%%%%%%%%%%%

\newpage
\appendix

\vskip 0.2in
\hrule
\startcontents[sections]
\printcontents[sections]{}{1}{\setcounter{tocdepth}{3}}
\vskip 0.2in
\hrule

\section{Implementation and Architectural Details}

\subsection{Detailed Implementation Setup}
\label{appendix:Detailed Implementation Setup}
\paragraph{Training and Evaluation Protocols}
During the training phase, we strictly adhere to the data processing protocols established for the SAM3~\citep{sam3} model. To maintain batch consistency across varying image scales, all input images are randomly resized to dimensions within the range of 480 to 1008 pixels utilizing a stride of 32, and subsequently padded to a unified spatial resolution of $1008 \times 1008$. In addition, standard data augmentation techniques, including random horizontal flipping and color jittering, are uniformly applied. During the inference phase, all images are processed using the standard $1008 \times 1008$ input resolution natively required by the SAM3 architecture. Furthermore, to ensure fairness in the evaluation, all experimental results are derived directly from the raw outputs of the model without the application of any post-processing techniques (\eg, PAMR~\citep{pamr} or DenseCRF~\citep{densecrf}).

\paragraph{Measurement of Inference Latency}
For a fair and rigorous evaluation of inference efficiency as reported in \Figure~\ref{fig:efficiency_comparison} and \Table~\ref{tab:ablation_study}, all inference latency measurements are conducted on a single NVIDIA RTX 4090 GPU. We report the per-image processing time configured with a batch size of 1. To eliminate hardware initialization and memory allocation overhead, we perform 50 warm-up iterations prior to measurement, and the final latency is calculated as the average execution time of the subsequent forward passes.

\begin{figure}[t]
    \centering
    \includegraphics[width=\linewidth]{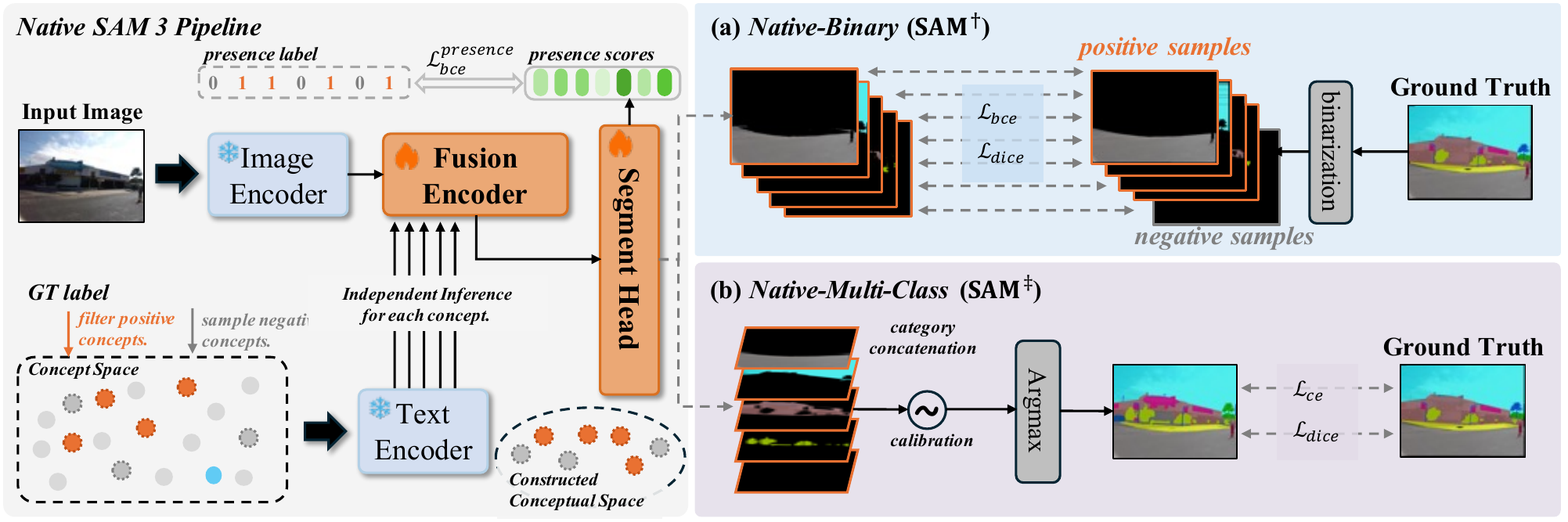}
    \caption{\textbf{Overview of two SAM3 baseline models for multi-class semantic segmentation}. We detail the construction of the concept space and the differences in training settings between the two baselines.}
    \label{fig:overview_baseline}
\end{figure}

\paragraph{Configurations for SAM3 Baselines}
To thoroughly evaluate our method, we fine-tune the pre-trained SAM3 model under two distinct paradigms. As shown in \Figure~\ref{fig:overview_baseline}, for both baselines, the image encoder and the text encoder are strictly frozen to preserve broad generic priors, while only the fusion encoder and the segmentation heads are fully optimized. For foreground categories, we directly utilize the original class names provided by the dataset, whereas a generic text prompt, \ie, ``background'', is employed to represent the background category. The detailed settings are as follows:

\textbf{I}) \textit{Native-Binary} (\textbf{SAM3$^\dagger$}): This baseline strictly adheres to the official strategy of SAM3. The fusion encoder processes each text prompt independently in a serial $\mathcal{O}(n)$ manner. The model is optimized using isolated binary prediction heads, supervised by a combination of the binary cross-entropy loss, dice loss, and presence loss for each individual concept. To construct the training data for this binary paradigm, we decouple the dense ground-truth annotations into independent binary masks based on the target categories present in the current image. Furthermore, we randomly introduce negative samples (\ie, unrepresented categories with completely empty ground-truth masks) during training to enhance the model's semantic discriminability and effectively suppress false positive predictions.

\textbf{II}) \textit{Native-Multi-Class} (\textbf{SAM3$^\ddagger$}): To adapt the native SAM3 architecture for a conventional multi-class setting, we concatenate the independently generated logits along the category dimension. A standard $\mathtt{softmax}$ activation is then applied to explicitly calibrate probabilities across the $N$ classes, followed by optimization via the cross-entropy loss and dice loss. To prevent Out-of-Memory (OOM) errors caused by processing an excessive number of categories simultaneously, we employ a dynamic prompt sampling strategy during training: we sample all ground-truth categories present in the current scene alongside a dynamically varying number of negative categories, ensuring a stable memory footprint per forward pass.

\paragraph{Hard-Coding Background Strategy Details}
In the component ablation study presented in \Table~\ref{tab:ablation_study}, we evaluated a hard-coding background strategy as a comparative baseline against our proposed dynamic background token ($\mathcal{T}_{bg}$). To implement this strategy without extensive manual annotation, we employed large language models (LLMs) to automatically generate a comprehensive, dataset-specific list of negative concepts (\eg, ``sky'', ``road'', and ``cloud'') that frequently appear in scenes but belong to the background class. Specifically, we utilized GPT-5~\citep{gpt-5} to generate $K$ diverse background concepts tailored to the domain of each benchmark. To prevent excessive semantic overlap with existing foreground categories, we applied the filtering pipeline of previous work~\citep{test-time-contrast} to these background concepts. First, we discarded visually uninformative and abstract concepts, such as ``image'' and ``view''. Subsequently, we used the CLIP~\citep{clip} model to compute the semantic similarity between the generated background concepts and the predefined foreground categories. Any background concept exhibiting a similarity score exceeding a threshold of 0.8 was removed. The specific prompts used are detailed below:

\prompt{
You are an expert in computer vision and semantic segmentation. I need you to generate \{$K$\} background or negative concept words that typically appear within the scenes of \{dataset name\} dataset:

\{dataset scene type, foreground categories, and additional related information\}.

These concepts should encompass a wide range of background elements, environmental descriptors, or non-target objects to ensure high diversity.

examples: ``asphalt road'', ``cloudy sky'', ``dense forest'' and ``bare soil''. 

All generated terms must be in lowercase English. You are required to output exclusively a Python list containing \{$K$\} strings, formatted similarly to [``sky'', ``road'', ``water''], and to omit any additional explanations or code block markers.
}

\subsection{Expanded Architectural Details of DCR-SAM}

To complement the methodological overview (\Sec\ref{method:dcr_sam}) in of DCR-SAM, we detail the micro-architectural designs and precise mathematical formulations of our proposed modules, ensuring complete structural transparency.

\paragraph{Vision-Conditioned Text Refinement (VCTR)}
In practice, the \textit{Vision-Conditioned Text Refinement (VCTR)} module is implemented using the standard $\mathtt{nn.MultiheadAttention}$ function provided by the $\mathtt{PyTorch}$ framework. Specifically, the text tokens $p_i \in \mathbb{R}^{l \times d}$ serves as the \textit{Query} ($Q$), while the globally contextualized visual representation $\mathcal{V}^* \in \mathbb{R}^{hw \times d}$ acts simultaneously as both the \textit{Key} ($K$) and the \textit{Value} ($V$). To ensure optimization stability, this process is naturally accompanied by standard Layer Normalization~\citep{layernorm} ($\mathrm{LN}$) and residual connections~\citep{resnet}. The entire module relies exclusively on highly optimized matrix operations, thereby introducing minimal computational latency to the overall inference pipeline (please see \Table~\ref{tab:ablation_study} for more details).

\paragraph{Vision-Guided Text Pooling (VGTP)}
Regarding the dimensional broadcasting in \Eq~\ref{eq:attention_score} and \Eq~\ref{eq:t_anchor}, the global visual embedding $\mathcal{V}_g \in \mathbb{R}^{1 \times d}$ is implicitly broadcasted along the category dimension $N$ to match the shape of the aligned text sequences. Crucially, the attention computation is strictly performed per class rather than being shared. The dot product between $\mathcal{V}_g$ and $\tilde{\mathcal{P}}$ yields an attention matrix $\mathcal{A} \in \mathbb{R}^{N \times l}$, where the softmax activation normalizes the weights strictly along the sequence length $l$ for each specific category. This ensures that the aggregated text anchors $\mathcal{T}_{anchor}$ capture class-specific discriminative cues without inter-class sequence interference.

\paragraph{Dense Competitive Head}
After obtaining the dense logits map $\mathcal{M}^{*} \in \mathbb{R}^{(N+1) \times hw}$, we subsequently reshape it back to the 2D spatial dimensions of $\mathbb{R}^{(N+1) \times h \times w}$. A bilinear upsampling algorithm~\citep{bilinear} is then applied to align the resolution with that of the original input image. Following this, an $\mathtt{argmax}$ operation is performed along the category dimension to produce the final segmentation masks.

\section{Additional Experimental Analysis and Discussion}

\subsection{Further Experimental Analysis}
\label{appendix:Further Experimental Analysis}

\paragraph{Impact of Concept Space Scale on Performance}
As observed in \Table~\ref{tab:main_res}, DCR-SAM achieves more pronounced performance gains on remote sensing benchmarks compared to natural scene benchmarks. We hypothesize that this phenomenon is fundamentally driven by the scale of the concept space: the massive number of fine-grained categories in natural scenes exacerbates feature space crowding and semantic competition. To validate this hypothesis, we conducted a progressive subsetting experiment on the highly complex ADE20K~\citep{ade20k} dataset ($N=150$). Specifically, we establish a base set comprising the Top-30 most frequent categories and progressively expand the target category count $N \in \{30, 60, 90, 150\}$. For subsets where $N < 150$, all pixels belonging to unselected categories are uniformly absorbed by our dynamic background token ($\mathcal{T}_{bg}$). Crucially, we isolate and track the performance of the identical Base-30 classes across all experimental configurations.

As shown in \Table~\ref{tab:category_scale_ade20k}, As the number of semantic categories increases ($30 \rightarrow 150$), the overall performance declines due to the introduction of difficult long-tail classes. Furthermore, the isolated \textbf{Base-30} class also suffers a substantial degradation, dropping from 79.4\% to 70.3\%. This indicates that even the most well-represented foundational categories are negatively impacted by the severe feature competition and representational crowding triggered by a larger concept space. Consequently, in remote sensing scenarios where the concept space is typically constrained (\eg, $N \le 10$), the dense competitive head can establish highly stable and non-overlapping semantic anchors. This effectively unlocks the full discriminative potential of the model architecture, thereby yielding superior performance gains.

\begin{table}[h]
    \centering
    \caption{\textbf{Impact} of the concept space scale on the ADE20K dataset.}
    \footnotesize
    \label{tab:category_scale_ade20k}
    \begin{tabularx}{\linewidth}{ccccc}
        \toprule
        \textbf{Target Category Count} & \textbf{Semantic Space Composition} & \textbf{Time.$\downarrow$} & \textbf{Overall mIoU (\%)} & \textbf{Base-30 mIoU (\%)} \\
        \midrule
        $N=30$  & Top-30 frequent classes only       & 236 & \bf 79.4 & \bf 79.4 \\
        $N=60$  & Base-30 + Next 30 frequent classes & 238 & 75.6\tiny{\textcolor[RGB]{191, 144, 0}{{$\downarrow$}3.8}}     & 77.5\tiny{\textcolor[RGB]{191, 144, 0}{{$\downarrow$}1.9}} \\
        $N=90$  & Base-30 + Next 60 frequent classes & 241 & 71.5\tiny{\textcolor[RGB]{191, 144, 0}{{$\downarrow$}7.9}}     & 73.9\tiny{\textcolor[RGB]{191, 144, 0}{{$\downarrow$}5.5}} \\
        $N=150$ & Full concept space                 & 244 & 68.7\tiny{\textcolor[RGB]{191, 144, 0}{{$\downarrow$}10.7}}    & 70.3\tiny{\textcolor[RGB]{191, 144, 0}{{$\downarrow$}9.1}} \\
        \bottomrule
    \end{tabularx}
\end{table}

\begin{figure}[t]
    \centering
    
    % --- 第一行：放置图片，使用 [c] 参数使其垂直居中对齐 ---
    \begin{minipage}[c]{0.5\textwidth}
        \centering
        % 替换为你的第一张图片文件名，可按需调整 width
        \includegraphics[width=\linewidth]{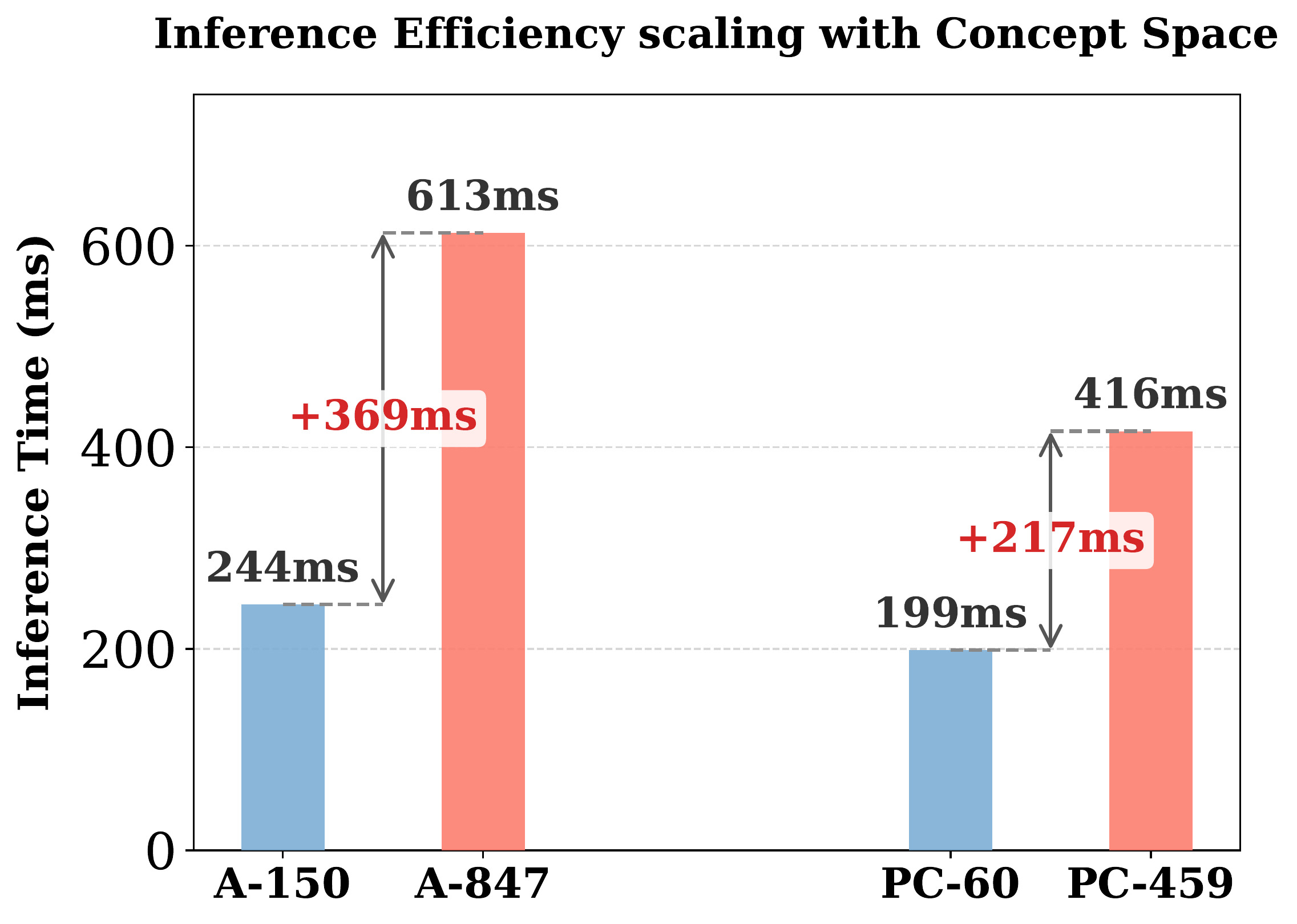}
    \end{minipage}\hfill
    \begin{minipage}[c]{0.45\textwidth}
        \centering
        % 替换为你的第二张图片文件名
        \includegraphics[width=\linewidth]{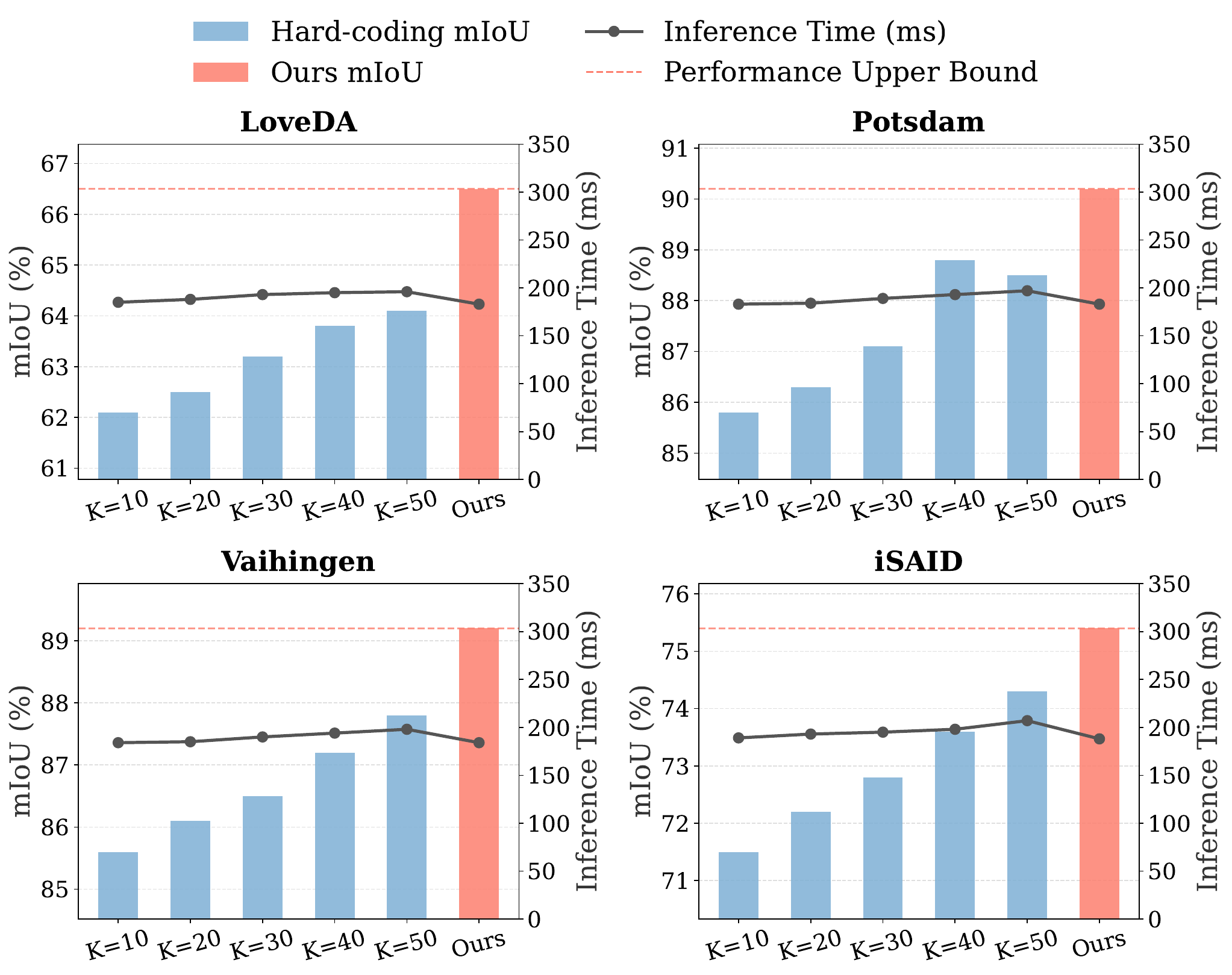}
    \end{minipage}
    
    \vspace{0.8em} % 调整图片与图注之间的垂直间距
    
    % --- 第二行：放置图注，使用 [t] 参数使其顶部（首行）对齐 ---
    \begin{minipage}[t]{0.5\textwidth}
        \caption{\textbf{Impact} of the concept space scale on inference efficiency.}
        \label{fig:efficiency_scaling}
    \end{minipage}\hfill
    \begin{minipage}[t]{0.45\textwidth}
        \caption{\textbf{Impact} of hard-coding background concept number $K$.}
        \label{fig:ablation_hardcoding_k}
    \end{minipage}
    
\end{figure}

\paragraph{Impact of Concept Space Scale on Inference Efficiency}
We further investigate the impact of the concept space scale on inference latency. Specifically, we evaluate the DCR-SAM framework on the ADE20K dataset with 150 categories (\textbf{A-150}) alongside its full version comprising 847 categories (\textbf{A-847}), as well as the PASCAL Context dataset with 59 categories (\textbf{PC-59}) and its full version comprising 459 categories (\textbf{PC-459}). As illustrated in \Figure~\ref{fig:efficiency_scaling}, expanding the target categories from 150 to 847 on ADE20K introduces a manageable latency overhead of $+369$\,ms (244\,ms $\rightarrow$ 613\,ms). Similarly, scaling from 59 to 459 categories on PASCAL Context incurs a mere $+217$\,ms increase (199\,ms $\rightarrow$ 416\,ms). Compared to the staggering 8038\,ms required by the native SAM3 on the \textbf{A-150} benchmark (please see \Table~\ref{tab:ablation_study}), these fractional increments are practically negligible. While our core cross-modal feature fusion process remains strictly category-agnostic, the lightweight matrix operations within the \textit{VCTR} module and the dense competitive head naturally introduce a sub-linear computational overhead as the number of target concepts $N$ grows. Nevertheless, achieving high-fidelity semantic segmentation for nearly a thousand categories in approximately 600\,ms thoroughly demonstrates the extreme efficiency and scalability of our architecture, effectively bypassing the severe $\mathcal{O}(N)$ complexity bottleneck inherent to the native SAM3 baseline.

\paragraph{Investigating Spatial-Preserving Pooling Strategies}
Our \textit{VGTP} module employs Global Average Pooling (GAP) on the visual features $\mathcal{V}^*$ to generate a global visual embedding $\mathcal{V}_g$ (\cf~\Eq\ref{eq:vgtp_global}). A reasonable concern is whether this GAP operation collapses fine-grained spatial cues, which are often deemed critical for resolving small objects in dense segmentation tasks. To investigate this, we conducted an ablation study comparing our GAP-based \textit{VGTP} against a Spatial-Preserving Attention (SPA)~\citep{spanet} variant. Specifically, we bypassed the GAP operation and flattened the spatial dimensions, computing a dense similarity matrix $\mathcal{S}$ directly between the aligned text sequence $\tilde{\mathcal{P}} \in \mathbb{R}^{N \times l \times d}$ and the full spatial visual features $\mathcal{V}^* \in \mathbb{R}^{hw \times d}$. The spatial-aware attention weights $\mathcal{A}_{spa} \in \mathbb{R}^{N \times l}$ are obtained by marginalizing this dense similarity matrix over the spatial dimensions:
\begin{equation}
    \mathcal{A}_{spa} = \mathtt{softmax}\left(\frac{1}{hw} \sum_{i=1}^{hw} \mathcal{S}_{:,:,i}\right),\quad \mathrm{where}\,\,S=\frac{\tilde{\mathcal{P}} (\mathcal{V}^*)^T}{\sqrt{d}},\,\, \mathcal{S} \in \mathbb{R}^{N \times l \times hw}.
\end{equation}
The final semantic anchors are subsequently aggregated using these spatial-aware weights, formulated as $\mathcal{T}_{anchor} = \sum_{i=1}^{l} (\mathcal{A}_{spa})_{:,i} \odot \tilde{\mathcal{P}}_{:,i}$.

As presented in \Table~\ref{tab:pooling_ablation}, the incorporation of the full spatial resolution in the SPA variant elevates the computational complexity of the attention mechanism. This introduces a slight inference latency and a concurrent performance decline (\eg, a 4.8\% mIoU drop on the ADE20K dataset). We argue that the retention of the full spatial resolution during the anchor generation phase compels the model to overfit to localized, noisy visual artifacts. In contrast, the GAP operation serves as an effective information bottleneck. It distills robust global representations from the features to reliably guide the aggregation of the text sequence, while delegating the high-fidelity spatial delineation to the final Dense Competitive Head.

\begin{table}[t]
    \centering
    \footnotesize
    \caption{\textbf{Ablation study} on the text sequence pooling strategy across four diverse datasets.}
    \label{tab:pooling_ablation}
    \begin{tabularx}{\linewidth}{lYYYYYYYY}
    \toprule
    \multirow{2}{*}{\textbf{Pooling Strategy}} & \multicolumn{2}{c}{\textbf{ADE20K}} & \multicolumn{2}{c}{\textbf{Cityscapes}} & \multicolumn{2}{c}{\textbf{LoveDA}} & \multicolumn{2}{c}{\textbf{Potsdam}} \\
    \cmidrule(lr){2-3} \cmidrule(lr){4-5} \cmidrule(lr){6-7} \cmidrule(lr){8-9}
     & \bf Time.$\downarrow$ & \bf mIoU & \bf Time.$\downarrow$ & \bf mIoU & \bf Time.$\downarrow$ & \bf mIoU & \bf Time.$\downarrow$ & \bf mIoU \\
    \midrule
    SPA (Spatial-Preserving) & 248 & 63.9 & 184 & 81.5 & 187 & 59.6 & 185 & 85.8 \\
    \rowcolor{green!5}
    \textbf{\textit{VGTP} (GAP-based)} & \textbf{244} & \textbf{68.7} & \textbf{183} & \textbf{89.1} & \textbf{183} & \textbf{66.5} & \textbf{183} & \textbf{90.2} \\
    \bottomrule
    \end{tabularx}
\end{table}

% \begin{wrapfigure}{r}{0.5\textwidth}
%     \includegraphics[width=\linewidth]{figures/ablation_hardcoding_k.pdf}
%     \captionof{figure}{\textbf{Impact} of hard-coding background concept number $K$.}
%     \label{fig:ablation_hardcoding_k}
% \end{wrapfigure}

\paragraph{Impact of Hard-Coding Background Concept Number}
To thoroughly investigate the efficacy and potential limitations of the hard-coding strategy, we conducted an ablation study varying the number of generated background concepts, denoted as $K$. Using the filtered background concepts (please see \Appendix~\ref{appendix:Detailed Implementation Setup}), we reported the overall mIoU and inference time on the remote sensing benchmarks by progressively scaling the hard-coding concept number, \ie, $K \in \{10, 20, 30, 40, 50\}$. During the inference phase, we appended these filtered concepts to the target category list to explicitly absorb non-target pixels. As shown in \Figure~\ref{fig:ablation_hardcoding_k}, due to the highly efficient, decoupled architecture of DCR-SAM, processing these additional concepts introduces only a marginal computational overhead rather than a severe latency bottleneck. However, the primary limitation emerges in the semantic performance, while scaling up $K$ initially improves background absorption, the performance gains rapidly saturate, even with a large set of explicitly designed negative concepts, the rigid textual anchors struggle to encompass all unconstrained background variations, resulting in suboptimal performance compared to our proposed dynamic background strategy ($\mathcal{T}_{bg}$), which consistently achieves superior performance, demonstrating that it provides a more elegant and adaptable solution for robust semantic masking without the necessity of exhaustive prompt engineering.

\paragraph{Compared with batched inference mode}
To further evaluate inference efficiency, we compare DCR-SAM with native SAM 3 under different batch sizes on ADE20K, which contains 150 categories. As shown in \Table~\ref{tab:batched_inference}, increasing the batch size from 1 to 50 reduces the inference latency of SAM 3 from 8038 ms to 5687 ms per image, but increases memory consumption from 9010 MB to 29080 MB. In contrast, DCR-SAM achieves a substantially lower latency of 244 ms with only 6318 MB of memory.

The limited acceleration of batched inference stems from the architecture of SAM 3, which involves backbone inference, feature fusion, and mask decoding. Although batching reduces redundant backbone computations by sharing visual features across categories, feature fusion and mask decoding must still be performed separately for each category, resulting in increasing computational overhead as the number of categories grows. In contrast, DCR-SAM performs backbone inference and feature fusion only once, with negligible additional overhead from the lightweight matrix operations in the VCTR module. These results demonstrate the superior computational efficiency and memory utilization of DCR-SAM in large-scale multi-class segmentation.

\begin{table}[h]
    \centering
    \footnotesize
    \caption{\textbf{Efficiency comparison} between SAM 3 and DCR-SAM across different batch sizes.}
    \label{tab:batched_inference}
    \begin{tabularx}{\linewidth}{lYYY}
    \toprule
    \textbf{Inference Strategy} &
    \textbf{Backbone Passes} &
    \textbf{Time (ms)$\downarrow$} &
    \textbf{Memory (MB)$\downarrow$} \\
    \midrule
    SAM 3 (Sequential, bs=1) & 150 & 8038 & 9010 \\
    SAM 3 (Batched, bs=15)   & 10  & 6392 & 12960 \\
    SAM 3 (Batched, bs=25)   & 6   & 6067 & 17488 \\
    SAM 3 (Batched, bs=30)   & 5   & 5966 & 19816 \\
    SAM 3 (Batched, bs=50)   & 3   & 5687 & 29080 \\
    \midrule
    \rowcolor{green!5}
    \textbf{\textit{DCR-SAM} (Ours)} &
    \textbf{1} &
    \textbf{244} &
    \textbf{6318} \\
    \bottomrule
    \end{tabularx}
\end{table}

\section{Expanded Qualitative Analysis}

\subsection{Cross-Modal Activation Map Visualizations}
\begin{figure}[!ht]
    \centering
    \includegraphics[width=0.6\linewidth]{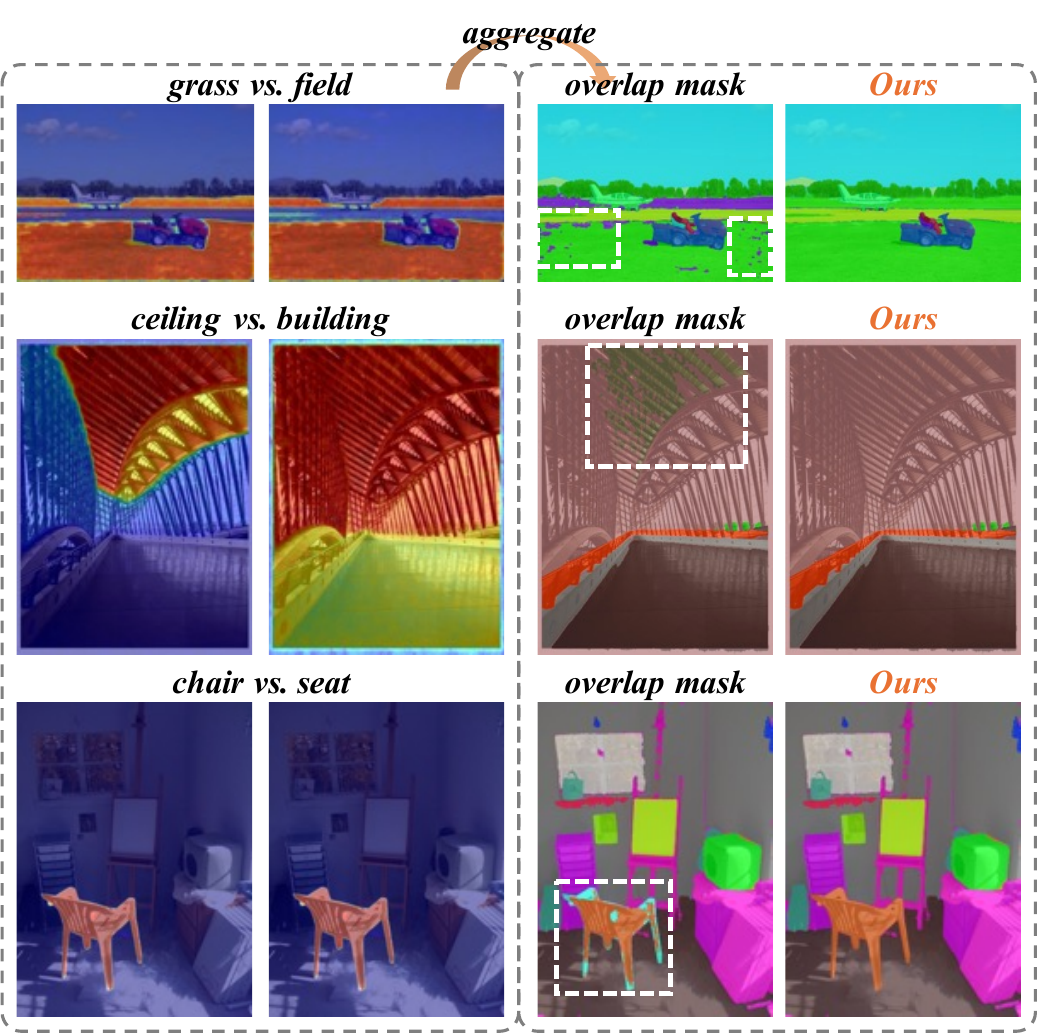}
    \caption{\textbf{Cross-modal semantic activation maps.} Aggregation in the original SAM3 leads to overlapping masks, whereas our method mitigates this issue through explicit inter-class competition.}
    \label{fig:additional_activation}
\end{figure}

To verify the effectiveness of our method in mitigating overlapping predictions, we conduct a visual analysis of the cross-modal activation maps. As illustrated in \Figure~\ref{fig:additional_activation}, it can be observed that directly aggregating the masks obtained from the independent inference of the native SAM3 results in overlapping predictions within the same regions. In contrast, our method effectively alleviates this issue through explicit inter-class competition.

\subsection{Extended Qualitative Comparisons}
\begin{figure}[!ht]
    \centering
    \includegraphics[width=0.8\linewidth]{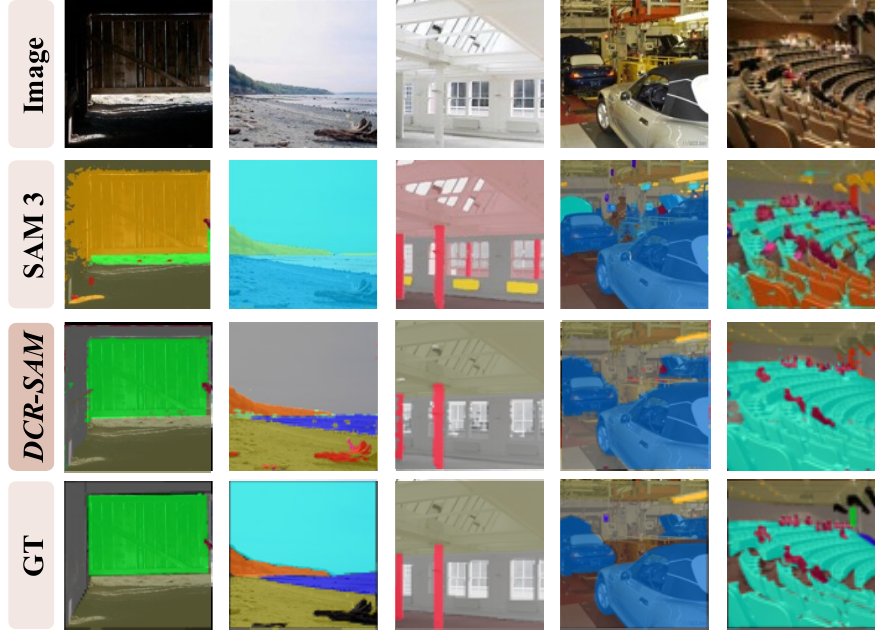}
    \caption{\textbf{Qualitative comparison} results on ADE20k dataset.}
    \label{fig:additional_ade20k}
\end{figure}

\begin{figure}[!ht]
    \centering
    \includegraphics[width=0.8\linewidth]{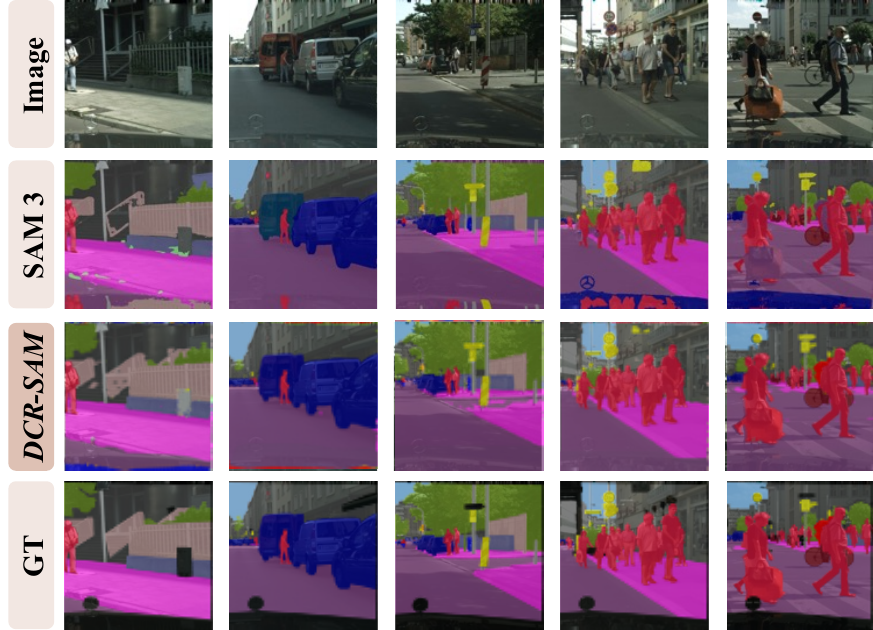}
    \caption{\textbf{Qualitative comparison} results on Cityscapes dataset.}
    \label{fig:additional_cityscapes}
\end{figure}

\begin{figure}[!ht]
    \centering
    \includegraphics[width=0.8\linewidth]{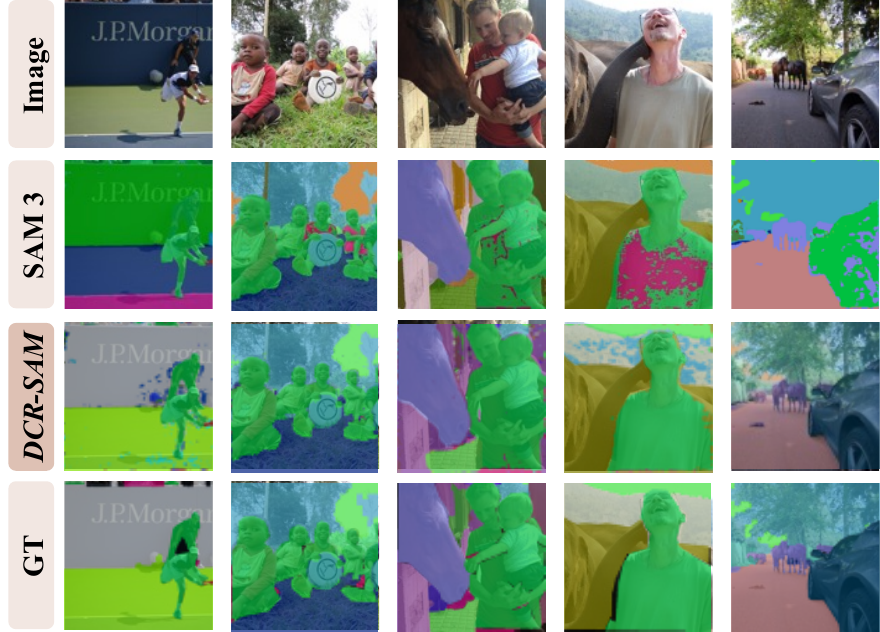}
    \caption{\textbf{Qualitative comparison} results on COCO-Stuff dataset.}
    \label{fig:additional_coco_stuff}
\end{figure}

\begin{figure}[!ht]
    \centering
    \includegraphics[width=0.8\linewidth]{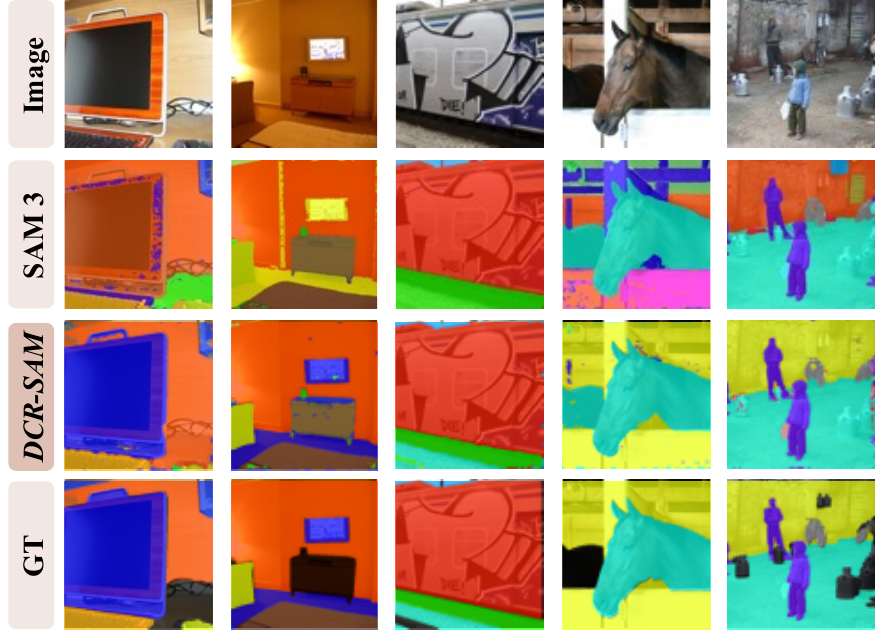}
    \caption{\textbf{Qualitative comparison} results on Context59 dataset.}
    \label{fig:additional_context59}
\end{figure}

\begin{figure}[!ht]
    \centering
    \includegraphics[width=0.8\linewidth]{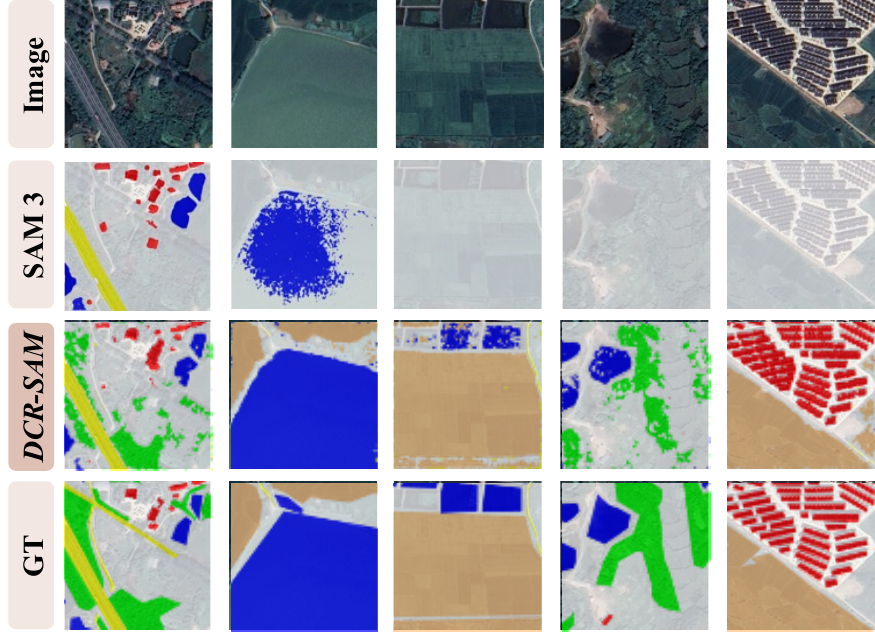}
    \caption{\textbf{Qualitative comparison} results on LoveDA dataset.}
    \label{fig:additional_loveda}
\end{figure}

\begin{figure}[!ht]
    \centering
    \includegraphics[width=0.8\linewidth]{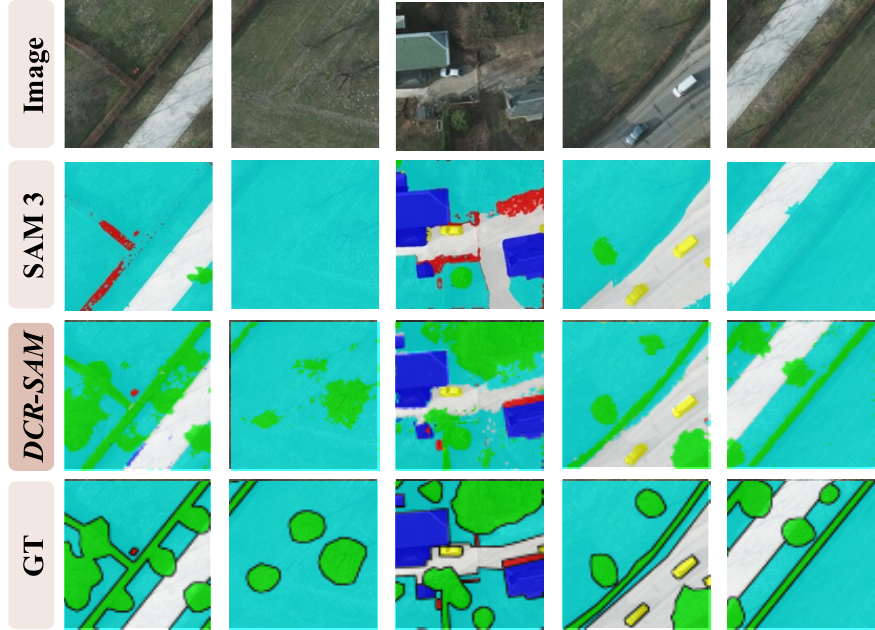}
    \caption{\textbf{Qualitative comparison} results on Potsdam dataset.}
    \label{fig:additional_potsdam}
\end{figure}

\begin{figure}[!ht]
    \centering
    \includegraphics[width=0.8\linewidth]{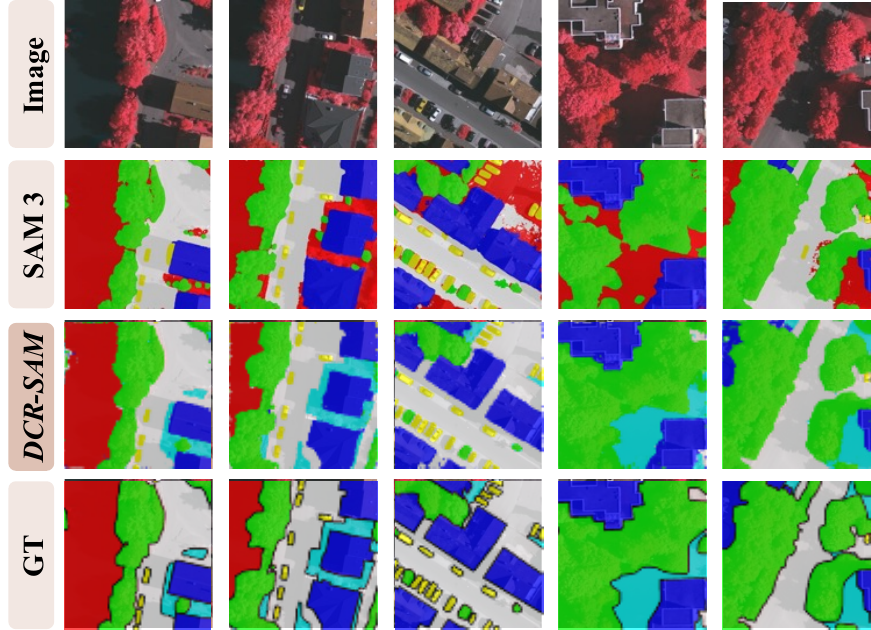}
    \caption{\textbf{Qualitative comparison} results on Vaihingen dataset.}
    \label{fig:additional_vaihingen}
\end{figure}

\begin{figure}[!ht]
    \centering
    \includegraphics[width=0.8\linewidth]{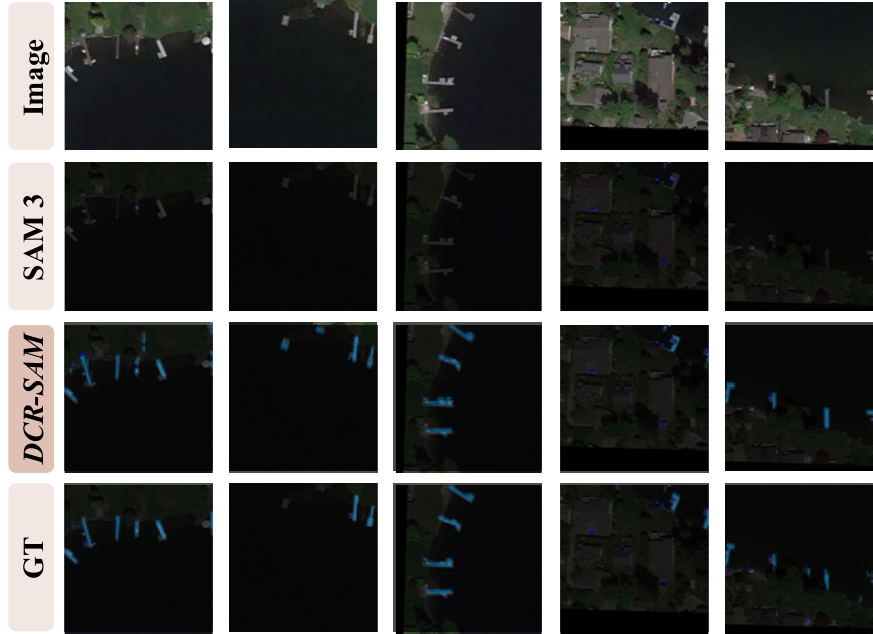}
    \caption{\textbf{Qualitative comparison} results on iSAID dataset.}
    \label{fig:additional_isaid}
\end{figure}

To further substantiate the qualitative superiority of our proposed framework, we present an extended set of visual comparisons across diverse natural scene and remote sensing benchmarks, as shown in \Figure~\ref{fig:additional_ade20k}-\ref{fig:additional_isaid}. These comprehensive visualizations demonstrate that the proposed method robustly delineates high-fidelity spatial boundaries while mitigating the severe semantic fragmentation prevalent in baseline models.

\end{document}